# Multimodal Marvels of Deep Learning in Medical Diagnosis: A Comprehensive Review of COVID-19 Detection

Md Shofiqul Islam*a,b*, Khondokar Fida Hasan*c,\**, Hasibul Hossain Shajeeb*d*, Humayan Kabir Rana*e*, Md Saifur Rahman*d*, Md Munirul Hasan*b*, AKM Azad*f*, Ibrahim Abdullah*g* and Mohammad Ali Moni*h,i,\*\**

*aInstitute for Intelligent Systems Research and Innovation (ISSRI), Deakin University, 75 Pigdons Rd, Warun Ponds, Geelong, 3216, Victoria, Australia*
*bUniversiti Malaysia Pahang Al-Sultan Abdullah(UMPSA), Kuantan, Pahang, 26300, Malaysia*
*cUniversity of New South Wales, UNSW, Canberra, 2601, ACT, Australia*
*dBangladesh University of Business and Technology, Mirpur-2, Dhaka, Bangladesh*
*eGreen University, Narayanganj, Dhaka, 1461, Bangladesh*
*fImam Mohammad Ibn Saud Islamic University, Al Thoumamah Rd, 11564, Riyadh, Saudi Arabia*
*gIslamic University, , Kushtia, 7600, Bangladesh*
*hThe University of Queensland, St. Lucia, Brisbane, 4072, QLD, Australia*
*iArtificial Intelligence and Cyber Futures Institute, Charles Stuart University, Panorama Ave, Bathurst, 2795, New South Wales, Australia*

**ABSTRACT**

This study presents a comprehensive review of the potential of multimodal deep learning (DL) in medical diagnosis, using COVID-19 as a case example. Motivated by the success of artificial intelligence applications during the COVID-19 pandemic, this research aims to uncover the capabilities of DL in disease screening, prediction, and classification, and to derive insights that enhance the resilience, sustainability, and inclusiveness of science, technology, and innovation systems. Adopting a systematic approach, we investigate the fundamental methodologies, data sources, preprocessing steps, and challenges encountered in various studies and implementations. We explore the architecture of deep learning models, emphasising their data-specific structures and underlying algorithms. Subsequently, we compare different deep learning strategies utilised in COVID-19 analysis, evaluating them based on methodology, data, performance, and prerequisites for future research. By examining diverse data types and diagnostic modalities, this research contributes to scientific understanding and knowledge of the multimodal application of DL and its effectiveness in diagnosis. We have implemented and analysed 11 deep learning models using COVID-19 image, text, and speech (ie, cough) data. Our analysis revealed that the MobileNet model achieved the highest accuracy of 99.97% for COVID-19 image data and 93.73% for speech data (i.e., cough). However, the BiGRU model demonstrated superior performance in COVID-19 text classification with an accuracy of 99.89%. The broader implications of this research suggest potential benefits for other domains and disciplines that could leverage deep learning techniques for image, text, and speech analysis.

## 1. Introduction

The advent of deep learning and machine learning technologies has revolutionised healthcare practice, generating greater efficiency and efficacy in a variety of medical tasks. In recent times, Artificial Intelligence performs well in the medical sector in diagnosing different diseases such as schizophrenia, ADHD Chen et al. (2019), epileptic seizures arrhythmia Islam et al. (2022c) Islam et al. (2023b). In particular, in the global fight against the COVID-19 pandemic, these technologies have been used extensively and diversely to surmount the myriad challenges that emerged Paka et al. (2021), Nugraheni et al. (2020). Deep learning (DL), in particular, has emerged as a powerful tool in the medical industry, providing decision support to clinicians rather than replacing human expertise Phillips-Wren and Ichalkaranje (2008). The promising accuracy and adaptability offered by deep learning technology almost ensures superior outcomes.

COVID-19, with common symptoms including fever, dry cough, myalgia, dizziness, and headache, has created significant detection difficulties due to the presence of asymptomatic cases Davenport and Kalakota (2019). To avoid these issues, text Paka et al. (2021), Nugraheni et al. (2020), speech Dastider et al. (2021) Lella and Pja (2022), and image-based Jain et al. (2021), Sitaula and Hossain (2021) methodologies have been widely used for the analysis and detection of COVID-19 Ardakani et al. (2020), Lella and Pja (2022), Paka et al. (2021). Different imaging modalities, such as CT and X-ray, are considered among the most successful strategies for the diagnosis of COVID-19 Ardakani et al. (2020) when accessible. Different imaging modalities, such as CT and X-ray, are believed to be among the most important methods for the diagnosis of COVID-19 Ardakani et al. (2020). Where available, CT scans

*Khondokar Fida Hasan
**Mohammad Ali Moni

✉ shafiqcseiu07@gmail.com (M.S. Islam); fida.hasan@unsw.edu.au (K.F. Hasan); hasibulhossainshajeeb@gmail.com (H.H. Shajeeb); humayan@cse.green.edu.bd (H.K. Rana); saifurs@gmail.com (M.S. Rahman); monirul.iiuc@gmail.com (M.M. Hasan); kazad@imamu.edu.sa (A. Azad); ibrahim@cse.iu.ac.bd (I. Abdullah); m.moni@uq.edu.au (M.A. Moni)
ORCID(s):

are preferred to X-rays due to their simplicity and three-dimensional pulmonary view Ardakani et al. (2020), Kim et al. (2020), although X-rays are more accessible and easier to read Ardakani et al. (2020).

Artificial Intelligence (AI) has emerged as a formidable ally in the fight against the novel coronavirus Kim et al. (2020). AI's applications in medical image processing have significantly improved the quality of diagnostic results while concurrently reducing the dependence on human intervention. In particular, deep learning and machine learning, both integral components of AI, have seen a surge in their application within medical scenarios. Emerging deep learning-based support mechanisms for the diagnosis of COVID-19 leverage a mix of text, speech, and image samples (CT and X-ray) Kim et al. (2020) Ardakani et al. (2020) Lella and Pja (2022) Paka et al. (2021). Some systems are based on pre-trained models of transfer learning Dourado Jr et al. (2019), while others utilise custom networks Ozturk et al. (2020).

This paper aims to explore the recent advances in deep learning-based COVID-19 diagnostic systems using data derived from medical imaging samples. We also delve into speech-based Dastider et al. (2021), Lella and Pja (2022) Sait et al. (2021), Pinkas et al. (2020), Pahar et al. (2021), Wang et al. (2020b) and text-based analyses Paka et al. (2021), Nugraheni et al. (2020), Li et al. (2021), Al-Laith and Alenezi (2021) using data from platforms such as Twitter. A comprehensive taxonomy is introduced to categorise the systems examined based on their usage of pre-trained models of deep transfer learning and custom deep learning techniques. We provide a critical evaluation of the most potent systems developed for COVID-19 diagnosis, taking into account diverse factors such as test data used, data splitting process, and measurement metrics Shuja et al. (2021). It is evident from our analysis that deep learning-based methods demonstrate superior precision and performance in image Khandokar et al. (2021), video Islam et al. (2020a), Islam et al. (2021a), speech Fan et al. (2020), and text Islam et al. (2021a), Khanday et al. (2020), Shofiqul et al. (2020), Islam et al. (2021b), Islam et al. (2022a) classifications.

## 1.1. Rational for the Study

Although there are several reviews and survey papers on machine- and deep-learning-based COVID-19 analysis, they differ greatly in their focus and depth. Several notable review papers based on deep learning are listed in Table 1 along with their respective contributions. For instance, Hasan et al. (2023) provides a systematic overview of deep learning applications utilising image data. However, its scope is limited to image-based COVID-19 analysis.

Similarly, a brief review article discusses audio, signal, speech, and language processing for COVID-19 Deshpande and Schuller (2020), while another presents a detailed survey of various deep learning methods and their performance, but does not address research gaps Subramanian et al. (2022). One paper explores different deep learning methods for COVID-19 analysis and their applications but lacks a methodological perspective Awassa et al. (2022).

A comprehensive review of COVID-19 analysis using deep learning is offered by Shorten et al. (2021), which outlines how neural networks feed information into deep neural networks, define learning tasks, and create new datasets and annotation protocols. However, it inadequately discusses the limitations of recent work, research gaps, and findings.

Additional reviews focus on COVID-19 analysis from audio data Lella and PJA (2021), machine learning-based approaches Syeda et al. (2021), and categorisation of current studies Islam et al. (2020a). However, they often overlook describing research gaps or providing a methodological perspective. Similarly, articles by Islam et al. (2020b), Manigandan et al. Manigandan et al. (2020), and Ardakani et al. (2020) offer analyses of COVID-19 but fail to elaborate on their methodologies or identify research gaps.

A review paper on COVID-19 analysis and prediction Salehi et al. (2020) succinctly analyses the results of seven models, but neglects to identify their shortcomings. Most existing surveys adopt a traditional approach, offering generalised explanations of methods and related works. Despite the abundance of research, only a limited number of review articles have been published for COVID-19 analysis of text, speech, and images Shuja et al. (2021).

In standard COVID-19 research surveys, a common pattern emerges a simplified comparative sample or an explanation of the associated study. The objectives and limitations of these reviews are summarised in Table 1.

Our survey paper diverges from this trend by providing an in-depth analysis of the taxonomy, architecture, and datasets of the COVID-19 studies. In addition, it offers an overview of data challenges, performance measures, and recent research references, including the architecture and equations of deep learning algorithms. In our critical review section, we dive into comparable studies, detailing their processes, data, number of class predictions, performance, and research gaps in all categories of deep learning-based COVID-19 analysis methods. This paper also provides a brief overview based on selected features.

Code and data available at https://github.com/shafiq-islam-cse/Multimodal-Marvels-of-Deep-Learning-Using-Image-Speech-and-Text-Review-of-COVID-19-Detection.

## 1.2. Contribution of the Study

Our main contribution to this review paper can be presented as follows:

- To present a systematic study on the state-of-the-art deep learning advances on speech, text, and medical imaging models of COVID-19 diagnostic systems with taxonomy, recent application, architecture, data sources, and prepossessing and challenges.

- State overall findings, performances, applications, and research gaps with related recent research reference, basic operation based on datatype, deep learning model with their completely illustrated architecture and equations, and relevant algorithm of COVID-19 analysis.

**Table 1**
Recent reviews on COVID-19 analysis.

| Author | Objective of the review | Limitation of the review |
|---|---|---|
| Shoeibi et al. (2024) | - Explores deep learning models to improve diagnostic accuracy and efficiency.<br>- Tackles challenges in interpretation and human error, with future directions in XAI and IoMT. | - Focuses on only experimental data and results, but does not cover novelty, finding, and applications<br>- Does not address all challenges with privacy concerns comprehensively. |
| Hasan et al. (2023) | Provides a systematic view of the application of deep learning method using image data. | - Lacks detailed methodology and evaluation processes.<br>- Limited to image-based COVID-19 analysis. |
| Deshpande and Schuller (2020) | Reviews deep learning methods using speech, text, and image data for COVID-19 analysis. | - Does not adequately cover all the advanced deep learning methods in depth.<br>- Lacks methodological gap analysis and potential recommendations. |
| Subramanian et al. (2022) | Illustrates different deep learning methods and their performance. | - This review focuses on the image domain (CT and X-ray) but does not cover the speech, cough, or text domains. |
| Awassa et al. (2022) | Reviews the application of deep learning methods for COVID-19 analysis. | - Lacks comprehensive presentation of multi-modal methods.<br>- Does not discuss challenges. |
| Shorten et al. (2021) | Presents methods and data annotation. | - Does not present recent works limitations and research gaps, results, and findings sufficiently. |
| Lella and PJA (2021) | Presents different methods with result analysis but focused on only speech-based data. | - Does not cover multi-modal data and its preprocessing |
| Syeda et al. (2021) | Better literature-based analysis on deep learning models performance. | - Does not cover the methodology in detail.<br>- Lacks coverage of limitations and privacy concerns of different deep learning methods. |
| Islam et al. (2020a) | Categorised method with a simple description. | - Does not present a sufficient analysis of findings, research scope, and limitations. |
| Islam et al. (2020b) | Focused on transmission, diagnosing, preventing, and imaging features for analysing COVID-19. | - Lack of describing methods in depth.<br>- Experimental analysis is not covered in the review |
| Manigandan et al. (2020) | Better analysis on the performance of ten deep learning models. | - Does not cover other things (novelty, findings, and potential application) except performances.<br>- Did not recommend discussion of the potentiality of the method. |
| Ardakani et al. (2020) | Offers an analysis of seven models with their results. | - It did not cover the application, potentiality, and research gaps in their narrow review.<br>- Not cover future works and recommendations of the best method. |
| Shuja et al. (2021) | Provides a general description of methods with a good approach. | - Did not cover challenges and privacy issues, but focused on the review of methods and algorithms. |

- Critical comparative analysis of all recent categories of deep learning-based COVID-19 analysis methods with advantages, disadvantages, limitations, future work, and concern about the most successful method.

- Analytically analyse the ten most advanced deep learning-based COVID-19 analysis methods manually and be concerned about the successful method.

- State the merits, demerits, challenges and limitations of the deep learning model and recommend the best tools in the deep learning-based COVID-19 analysis.

## 1.3. Novelty claim of the study

Existing research on using deep learning to diagnose COVID-19 is often broken down into reviews that focus on different aspects without a unified framework. We fundamentally claim that this research breaks this mould by giving an exceptionally thorough and well-organised analysis. We would like to pinpoint three sides that are novel in our review works in this paper;

*First*, we diverge from the summaries commonly found in other works by offering a detailed taxonomy that creates a clear and organized knowledge base for the field. This structure is enhanced by examining the latest applications in diverse domains such as speech, text, and medical imaging. We ensure unparalleled comprehensiveness by presenting data sources, in-depth preprocessing, applications, and the advancement of methods.

*Second*, we delve deeper than simply summarising model performance. We conduct a deep dive into the architecture of these models, unveiling the underlying methodologies and technical intricacies – a crucial aspect often missing from existing literature.

*Third*, by performing an in-depth experimental analysis of the ten most advanced methods, this analysis goes beyond what other studies offer, extracting the secrets behind these cutting-edge models and providing invaluable insights into their inner workings.

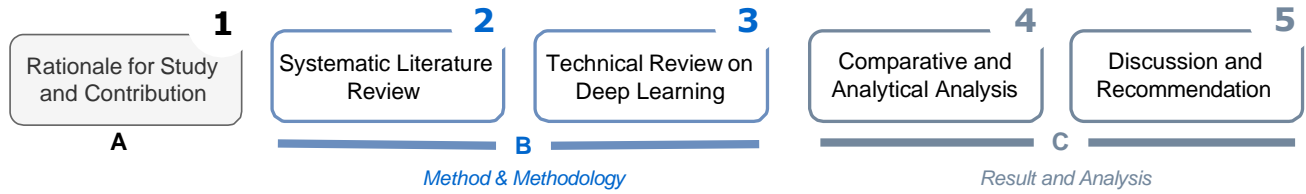

**Figure 1:** A high-level view of the organisation of the paper.

*Fourth,* we have provided a comprehensive recommendation and potential future work area for the best method after analyzing the outcomes, limitations, and challenges of various deep learning models.

Overall, this research offers a groundbreaking approach. It goes beyond the existing literature by providing a unifying framework, unveiling technical secrets, and translating the findings into practical applications. We don't just present findings; we provide a balanced evaluation of deep learning models, acknowledging their merits, limitations, and challenges.

### 1.4. Organisation of the Paper

At a high level, this paper can be seen as a three-step development as shown in Figure 1. All sections of this paper can be found as follows.

**(A.)**

**Section 1** presents the rationale, contribution, and organization of the paper.

**(B.)**

**Section 2** discusses the systematic literature review methodology, including a taxonomy of deep learning-based COVID-19 analysis approaches, data for COVID-19 analysis, and pre-processing tools.

**Section 3** covers the methodologies used for COVID-19 detection and analysis using deep learning techniques.

**(C.)**

**Section 4** presents the comparative and analytical result analysis, including recent methods, experimental setup, and overall performance.

**Section 5** addresses the discussion and recommendations, highlighting the novelty and limitations of the study.

Finally, **Section 6** provides the concluding remarks and future prospective work of the paper.

For the convenience of the reader, a list of abbreviations used throughout the paper is given in Table 2.

## 2. Systematic literature review methodology

This section presents the methodology followed to conduct the systematic literature review. Our article selection process follows the following rules:

**Table 2**
List of acronyms used in the paper

| Acronym | Meaning |
|---------|---------|
| AI | Artificial Intelligence |
| Att | Attention-based Neural Network |
| BOW | Bag Of Words |
| BiGRU | Bidirectional Gated Recurrent Unit |
| BiLSTM | Bidirectional Long Short Term Memory |
| Caps | Neural Network |
| CL | Computational Linguistic |
| CNN | Convolutional Neural network |
| COVID-19 | Corona Virus Disease 19 |
| CT | Computerised Tomography |
| DL | Deep learning |
| DNN | Deep Neural Network |
| FFT | Fast Fourier Transform |
| FFNN | Feed Forward Neural Network |
| FT | Fourier Transform |
| GAN | Generative Adversarial Network |
| GCN | Graph Convolutional Neural Network |
| GGO | GroundGlass Opacity |
| GRU | Gated Recurrent Unit |
| HDL | Hybrid Deep Learning |
| LSTM | Long Short Term Memory |
| MFC | Mel-frequency cepstrum |
| ML | Machine learning |
| MLP | Multi-Layer Perception |
| MRI | Magnetic Resonance Imaging |
| NN | Neural Network |
| NLP | Natural Language Processing |
| NPI | National Provider Identifier |
| RNN | Recurrent Neural Network |
| ROC | Receiver Operating Characteristic |
| SFT | Short Fourier Transform |
| SVM | Suppor Vector Machine |
| WHO | World Health Care Organisation |

1. Proper keyword selection in searching: We used some keywords like "COVID-19", "Tweet", "Cough", "X-ray and CT" "Hybrid" and "Deep Learning". Those keywords help to find the most relevant research article for COVID-19 analysis.

2. Relevant article selection: Focused on the most recent relevant methods with their application field in COVID-19 analysis.

3. Data source selection: We considered three types of data (Text, Cough, and Image) in this review.

4. Current deep learning method selection: In our article selection process, we consider the most recent and highly performed methods to analyse COVID-19, like

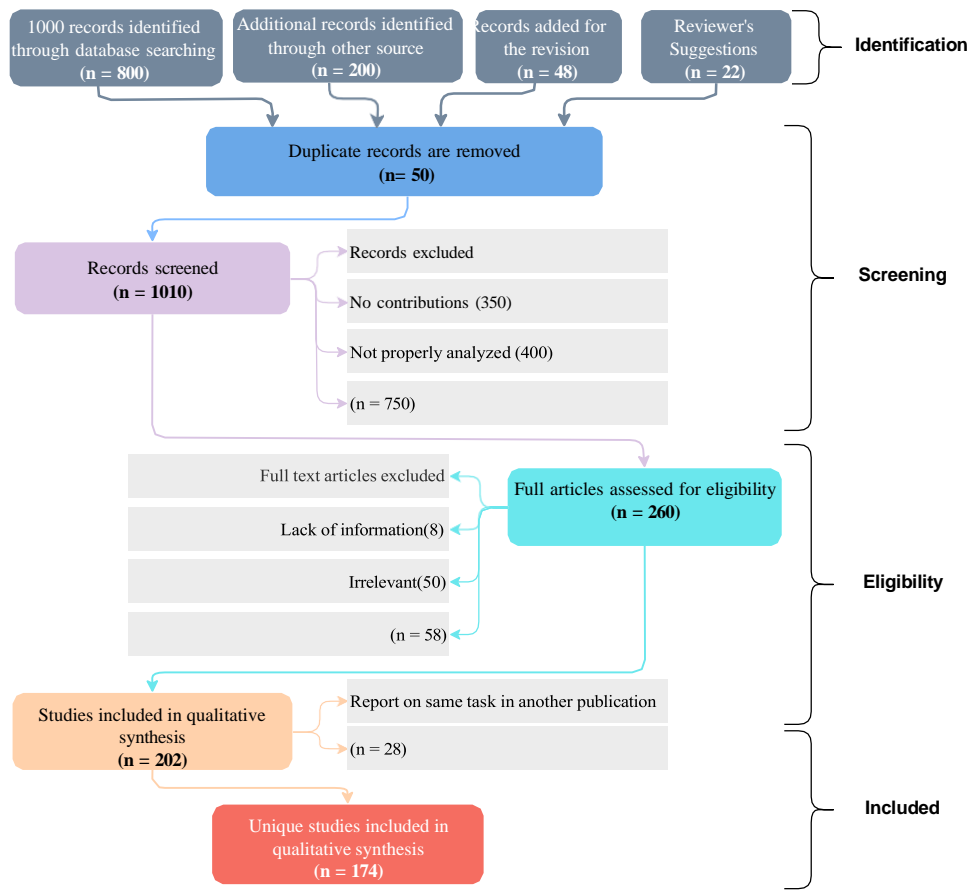

**Figure 2:** Flow diagram of literature search for including studies in our systematic review.

transfer learning, Attention, GAN, Capsule, RNN, and Hybrid method.

5. Time range selection: We focused only on the last three years' methods that are suitable to analyse COVID-19.

6. Quality selection: Article quality is defined by the validated production of data, good Journal or publisher (IEEE, Elsevier, SAGE, Wiley and Tralor & Francis and Springer)

7. Proper findings and applicability analysis: We also consider the relevant applicability and findings of the methods in analysing COVID-19 with text, image, and sound data.

Figure 2 shows a flow diagram of the literature screening process. This flow chart highlights the inclusion and exclusion of papers at each stage. All the papers are selected from different prestigious journals. The articles in this collection are the result of an online database search. Some of the online databases used include Scopus, IEEE Access and Explore, Springer, ACM library, Elsevier, SAGE, MDPI, Taylor and Francis, Wiley, Peerj, JSTOR, DBLP, and DOAJ. At the studying time, some papers are not considered for some cases that are mentioned in Figure 2 below. The final paper selection is improved at the screening and eligibility levels. Our search terms for the analysis yielded a total of 1010 relevant publications at first. We removed 700 references from these publications and considered the remaining 260. We excluded 58 studies based on the criteria given above (e.g. systematic reviews and review reports) and only reviewed full-text articles that examined the analysis in patients with COVID-19. Finally, we read the complete text of the remaining 202 papers and then eliminated 28 research studies since they were assigned the same task in another publication. In total, 174 publications met the inclusion criteria in our review analysis. We also have some recent papers in the relevant field, Table 3 presents the year-wise distribution of the reviewed article in this article for different COVID-19 domains.

## 2.1. Taxonomy of deep learning-based COVID-19 analysis approach

Many methods, approaches, terms, and techniques have been presented in scholarly articles related to COVID-19. However, we believe that a taxonomy should assist in understanding and developing ideas; thus, an overall taxonomy of COVID-19 analysis approaches is discussed in this section with a figure. A complete overview of deep learning approaches used for COVID-19 detection, analysis, and classification is shown in Figure 3a. COVID-19 can be analysed with the traditional approach (Statistical or Manual), traditional machine learning (SVM) Kohmer et al. (2020), Ismael and Şengür (2021), Random forest Wu et al. (2020b)) and the deep learning method. Statistical and manual approaches are normally used in the text-based analysis of COVID-19.

**Table 3**
Year-wise distribution of reviewed article in this article with some references

| Year | Total article | Methods (Covid Image) | Methods (Covid Cough) | Methods (Covid Text) |
|---|---|---|---|---|
| 2024 | 8 | Vision Transfer Mezina and Burget (2024). | CNN Chen et al. (2024) and Various. | Deep ensemble Muaad et al. (2024). |
| 2023 | 20 | ResNet Choudhary et al. (2023) Islam et al. (2023a), HR Net Al-Waisy et al. (2023), and Light CNN Khan et al. (2023). | Dense and convolution Ulukaya et al. (2023), VGGish, YAM-NET, and L3-Net Campana et al. (2023) and D-Cov19Net Chatterjee et al. (2023) | Transfer BERT Contreras Hernández et al. (2023), Lexicon-based deep learning Ahmed et al. (2023), RNN MLP CNN Paul et al. (2023), and BERT Lexicon Qorib et al. (2023). |
| 2022 | 25 | ResNet Chakraborty et al. (2022), Transfer Model Aggarwal et al. (2022), DenseNet Islam et al. (2022b), 2D-3D CNN Oda et al. (2022) and Reinforcement learning Allioui et al. (2022) | ML, DL Aly et al. (2022), DCNN Lella and Pja (2022) and ResNet Pahar et al. (2022) | CNN-LSTM Biswas and Dash (2022) and BERT DL Malla and Alphonse (2022). |
| 2021 | 39 | InceptionV3 XceptionNet Jain et al. (2021), Attention Zhang et al. (2021a) and VGG Sitaula and Hossain (2021), DNN Ozyurt et al. (2021), VGG16 and ResNet50 Zebin and Rezvy (2021), CNN Islam et al. (2020a) Ahsan et al. (2021), Ismael and Şengür (2021), FGCN Wang et al. (2021) and Capsule Network Afshar et al. (2020) Mobiny et al. (2020). | Inception Sait et al. (2021), LSTM ResNet Pahar et al. (2021) and CNN-LSTM Dastider et al. (2021) | AttentionBiLSTM Paka et al. (2021), BiLSTM Liang et al. (2021), AttentionGCNChe et al. (2021), and CNN-LSTM Islam et al. (2021b). |
| 2020 | 59 | Capsule Afshar et al. (2020), Mobiny et al. (2020) and CNN LSTM Islam et al. (2020a). | Attention-BiGRU Wang et al. (2020b), Attention GRUPinkas et al. (2020), VGG Brown et al. (2020), ResNet Laguarta et al. (2020b), CNN-LSTM Dastider et al. (2021). | CNN-LSTM Parashar et al. (2020), Attention BiLSTM-CRF Zhang et al. (2020), and GRU Raamkumar et al. (2020). |

The traditional machine learning algorithm is widely used for classification purposes and is popular for textual analysis. Our main focus is a deep learning-based COVID-19 analysis. However, the deep learning-based approach is classified into supervised, unsupervised, and semi-supervised categories. The most commonly used methods are CNN Sohrabi et al. (2020), Chakraborty et al. (2022), Bhattacharyya et al. (2022), Islam et al. (2020a), RNN Raamkumar et al. (2020) and hybrid Islam et al. (2020a), Shah et al. (2021), Islam et al. (2020b). CNN-based COVID-19 diagnosis and prediction is the most popular. The CNN-based classifier has some types of operation, such as attention, capsule, shallow, deep Lella and Pja (2022), Wu et al. (2020a), dynamic, deep CNN Ardakani et al. (2020), Ozyurt et al. (2021), and multilayer hybrid CNN Laguarta et al. (2020b), Jain et al. (2021). In the cased of RNN Imran et al. (2020), Hassan et al. (2020), Imran et al. (2020), it has two evolutions named LSTM Zhang et al. (2020) and GRU Raamkumar et al. (2020). Both LSTM Islam et al. (2020a) and GRU Paka et al. (2021) ResNet Al-Waisy et al. (2023) Islam et al. (2023a), Light CNN Khan et al. (2023)

Dense and convolution Ulukaya et al. (2023), Transfer BERT Contreras Hernández et al. (2023) Qorib et al. (2023), RNN MLP CNN Paul et al. (2023) can be used with both direction-based operations Passricha and Aggarwal (2020). Capsule network with RNN also gives good outcomes (Capsule Network) Islam et al. (2020a), Prabha and Rathipriya (2020), Malla and Alphonse (2021), Mobiny et al. (2020), Afshar et al. (2020). GAN Bar-El et al. (2021), Fan et al. (2020), Elzeki et al. (2021), Goel et al. (2021), Quilodrán-Casas et al. (2022), Zhu et al. (2020), Fan et al. (2020) model performs good than other. The attention mechanism Sitaula and Hossain (2021), Zhang et al. (2021a), Paka et al. (2021), Zhang et al. (2020), Wang et al. (2020b), Pinkas et al. (2020) sometimes gets great results. Graph Convolutional Network (GCN) Yu et al. (2021), Kapoor et al. (2020), Che et al. (2021), Wang et al. (2021), Laguarta et al. (2020a), Liang et al. (2021) also used for COVID-19 detection. Figure 3b shows the source and amount of data for the COVID-19 analysis.

By studying basic depictions, deep learning strategies can clarify complicated problems. With the ability to learn exact interpretations and the property of labelled training data, several layers are used sequentially. Deep learning algorithms Ardakani et al. (2020) are commonly employed in medical systems such as bio-medicine Kong et al. (2020), smart healthcare Esteva et al. (2019), drug development Chen et al. (2018), medical image recognition Kim et al. (2020), etc. It is also widely used in the automatic diagnosis of patients with COVID-19. A massive convolution neural network Ardakani et al. (2020) is equipped with lots of pretrained models that are being used in transfer learning (CNN Ahsan et al. (2021), Wu et al. (2020a)). There are some pretrained models which are being using for COVID-19 diagnosis such as AlexNet Ardakani et al. (2020), GoogleNet Ardakani et al. (2020), SqueezeNet Ardakani et al. (2020), different versions of Visual Geometry Group (VGG) Zebin and Rezvy (2021), Hemdan et al. (2020), diverse kinds of ResNet Wu et al. Wu et al. (2020b), Xu et al. (2020), Xception Ardakani et al. (2020), Hemdan et al. (2020), different forms of inception Jain et al. (2021), Horry et al. (2020), Hemdan et al. (2020), diverse types of MobileNet Ardakani et al. (2020), DenseNet Hemdan et al. (2020), encoder-decoder Amyar et al. (2020), capsule network Afshar et al. (2020), ResNet Al-Waisy et al. (2023) Islam et al. (2023a), Light CNN Khan et al. (2023) Dense and convolution Ulukaya et al. (2023), Transfer BERT Contreras Hernández et al. (2023), Qorib et al. (2023), RNN MLP CNN, Paul et al. (2023) etc.

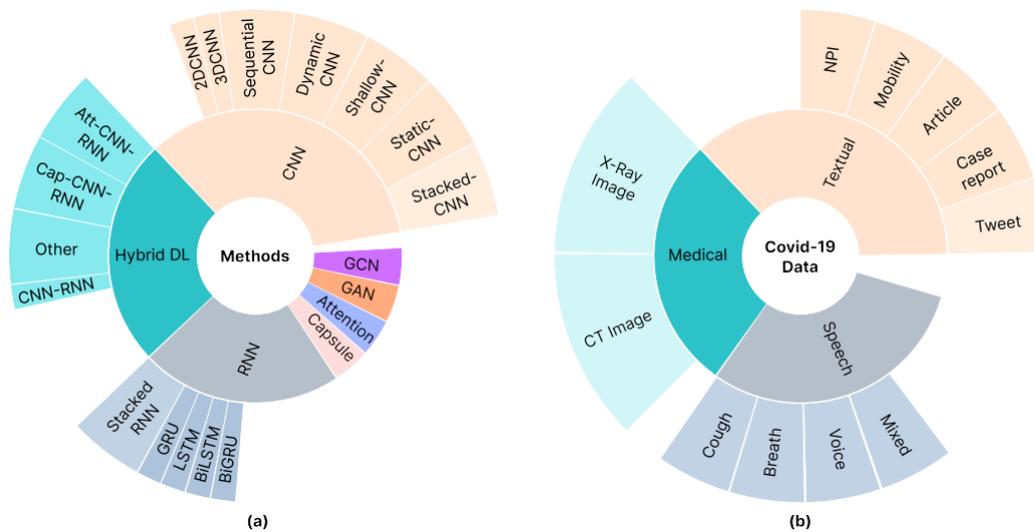

**Figure 3:** (a) Graph of COVID-19 analysis using deep learning method, (b) COVID-19 data source overview.

## 2.2. Data for COVID-19 Analysis

This section goes into relevant data for COVID-19 analysis. Many common data sets can be used in COVID-19 analysis for COVID-19 and polarity detection. This review article classifies deep learning data into three categories: medical, textual, and speech. Medical data are two types of CT scans and X-ray images from medical sources. Textual data for deep learning are collected from social networks or clinical laboratories. On the other hand, speech data can be the sound of cough and breath. Figure 3b summarises the data sets, including the type of data set and its amount. And Table 4 presents the publicly available data and its source.

### 2.2.1. Text-based data for COVID-19 analysis

Medical and social media data are used in the text-based analysis of COVID-19. There are many open sources of textual data for COVID-19 analysis using natural language processing. The National Pandemic Initiative (NPI) provides a lot of data to combat the COVID-19 epidemic. Different medical and academic universities, such as Oxford University and NPI, launched a variety of textual data for analysis. Data are compiled from the website, news stories, and official press releases. On the team website, data are accessible in various formats and interfaces. On the organisation's website, Kaggle, and GitHub archive, data is accessed in a variety of formats as well as interfaces. Travel cost, system development, and region policies control data collection for COVID-19 analysis Shuja et al. (2021).

### 2.2.2. Speech-based data for COVID-19 detection and analysis

Speech-based data are three types: cough, breath, and voice Shuja et al. (2021). Recorded speech of COVID-19 patients is used to analyse COVID-19 patient recognition. Speech The speech (audio) data sets aid in COVID-19 detection and identification in three ways. The sounds of cough, for example, can help detect a COVID-19-infected case following the application of ML techniques Deshpande and Schuller (2020). Second, the breathing rate can be detected by expression, resulting in COVID-19 screening Greenhalgh et al. (2020). Finally, expression-based stress recognition methods can classify people with signs of mental health problems and the severity of COVID-19 symptoms. Many of these methods require intensive data collection activities. Smartphones may be used to allow these speech-based COVID-19 diagnostic techniques. GitHub, Kaggle, and Figshare are the most used and public sources of speech data for the detection and analysis of COVID-19. Chest ultrasound data are more prevalent in COVID-19 detection and analysis. Chest ultrasound data has some advantages: availability, low cost, and easy use in the non-invasive technique. However, these data need to be more pre-processed

**Table 4**
Data source (Public).

| Type | Amount | Class | Source |
|------|--------|-------|--------|
| Xray | 21,165 | 3616 COVID-19 positive, 10,192 Normal, 6012 Non-COVID and 1345 Viral Pneumonia. | https://www.kaggle.com/datasets/tawsifurrahman/covid19-radiography-database |
| Xray | 6216 | 1341 and 4875 | https://www.kaggle.com/datasets/paultimothymooney/chest-xray-pneumonia |
| Xray | 325 | COVID-19 and non-COVID-19 | https://github.com/ieee8023/covid-chestxray-dataset |
| CT Xray | 1124 | 403 COVID and 721 Normal | https://github.com/M4rc-C/OPENCovidGAN |
| Xray | N/A | COVID and Normal | https://www.kaggle.com/competitions/rsna-pneumonia-detection-challenge/data |
| Xray | N/A | COVID and Normal | https://github.com/agchung/Actualmed-COVID-chestxray-dataset |
| Xray | 232 | 120 COVID and 112 Normal CXR | https://twitter.com/chestimaging |
| Cough | 121 | 73 COVID and 48 Normal | https://www.ncbi.nlm.nih.gov/pmc/articles/PMC8320312 |
| Cough | 16 | 9 COVID and 7 Normal | https://github.com/virufy/virufy-data |
| Cough | 1000+ | N/A | https://github.com/iiscleap/Coswara-Data |
| Cough | 28,00 | N/A | https://www.kaggle.com/datasets/andrewmvd/covid19-cough-audio-classification |
| Cough | 155 | Covid 43 and Normal 112 | https://www.kaggle.com/datasets/neptuneapps/spartacov-detect-covid-from-cough |
| Cough | 1400 | Covid 1207, Normal 150, Other 43 | https://www.kaggle.com/datasets/pranaynandan63/covid-19-cough-sounds |
| Cough | 171 | Covid 24 and Normal 147 | https://www.kaggle.com/datasets/isodscompetitions/covid-cough-audio |
| Text | 44,955 | 7955 Neutral, 20000 Positive and 1700 Negative | https://www.kaggle.com/code/melisaademiiir/covid19-tweets-sentiment-analysis-nlp/data |
| Text | 40,00 | Positive 17250 Negative 15150 Neutral 7600 | https://www.kaggle.com/datasets/datatattle/covid-19-nlp-text-classification |

to avoid noise that sometimes leads to irrelevant information prediction.

### 2.2.3. Image-based data for COVID-19 detection and analysis

There are many sources of image-based data. GitHub and Kaggle are the best options for finding open sources of image-based data for analysis. Typically, images are of two types: X-ray and CT scan images. Both types have specific features, advantages, and disadvantages in research. X-ray image data have some advantages: Good to analyse, available. However, X-ray image-based data have some drawbacks, such as not being able to detect pulmonary embolism and inadequate pavement. Cannot handle loss GGO density-based image. X-ray image data have some advantages: low cost comparatively, high sensitivity, and good re-productivity. However, X-ray image-based data have some disadvantages, such as less availability and the need for several time scans to ensure availability. A well-known technique is medical image processing, which could help the COVID-19 analysis.

Several papers also published biometric analyses of COVID-19-related experimental works. The Allen Institute for AI and other partners launched the CORD-19 project to gather articles on COVID-19 studies Wang et al. (2020a). Researchers used Association Rule Text Mining (ARTM) and Information Cartography strategies for the same data collection. Researchers at Berkeley Lab have created an online search platform for the collection of data from scientific publications on COVID-19. Data collection comprises many academic data sets, including Wang et al. (2020a) Lite COVID and Elsevier's Novel Coronavirus Information Centre. NPI is a series of various government steps to tackle the COVID-19 pandemic. NPI data sets are essential for the research and interpretation of COVID-19 transmissions. NPI data sets are critical for analysing COVID-19 transmissions and analysing the impact of NPI on COVID-19 cases (infections, deaths, etc.). A team of Oxford University scholars and students systematically compiled freely available data from all over the world into a Stringency Index Hale et al. (2020). Most of this image data is listed in a research article during the COVID-19 era Shuja et al. (2021).

In this subsection, here, subfigures a,b c of Figure 4 are added to present the overview of the data set used in each model for the analysis of COVID-19 using text, speech and image, respectively. The X-axis of this figure indicates several data used in the model, and the colour bar represents a class of disease or opinion. A colour label is given on the right side of this figure for each class.

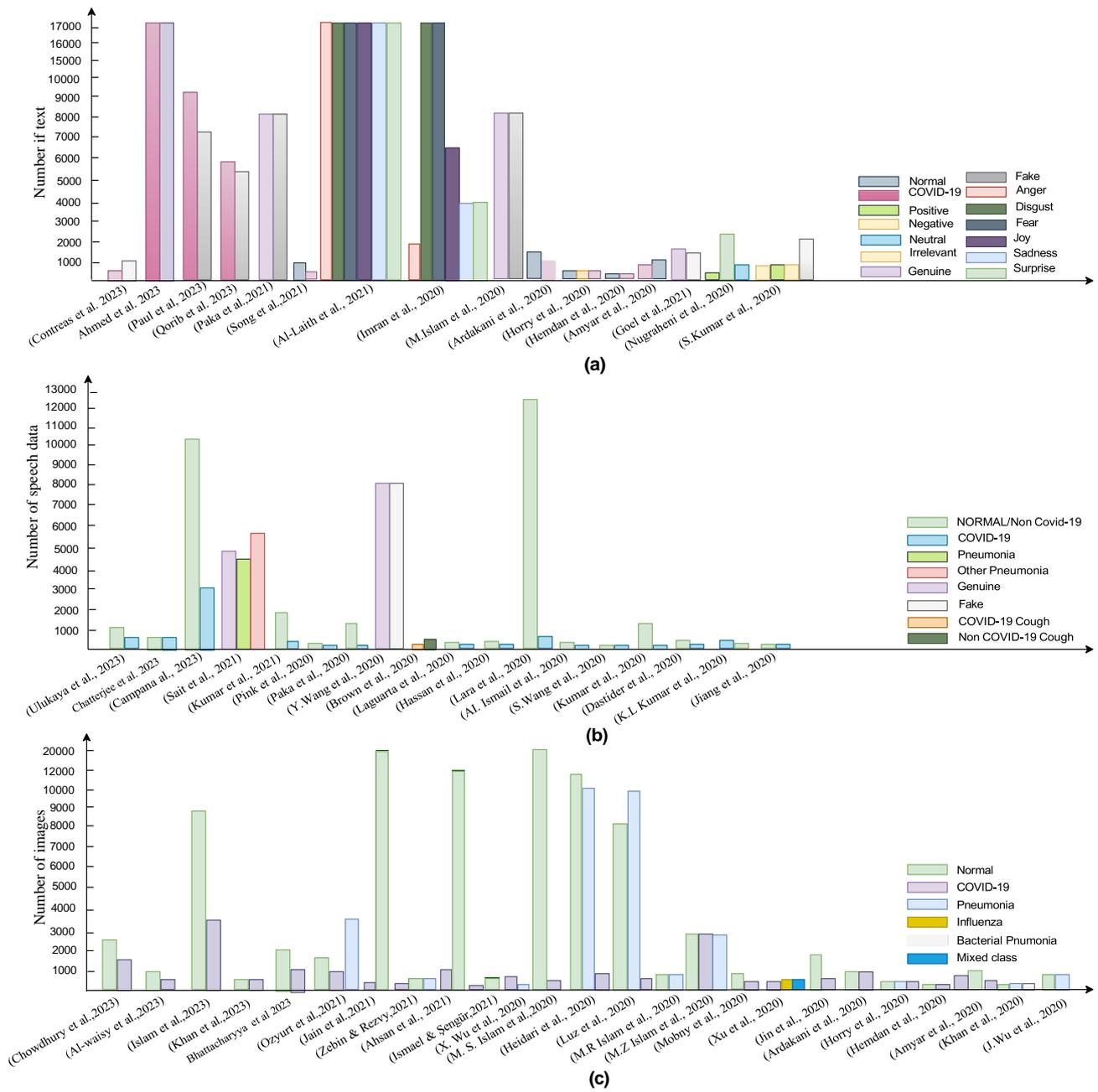

**Figure 4:** (a) The amount of text used for each model for COVID-19 analysis using text, (b) The amount of speech data used in each model for COVID-19 analysis using audio, (c) The amount of image data used in each model for covid-19 analysis using images.

In Figure 5, we presented the amount of analysis of COVID-19-based applications by type with percentage.

### 2.2.4. Common challenges in analysing COVID-19 data

This study categorised deep learning-based works into three types according to the data types: text, speech, and image-based approaches. This section discusses open-source data and challenges for COVID-19 analysis and prediction. Artificial intelligence and deep learning methods are often open-source and introduced as libraries, including packages in computer programming language-enabled development platforms. As a result, the emphasis shifts to data transparency and accessibility. Due to the novel COVID-19 epidemic, new data sets must be developed, managed, hosted, and benchmarked.

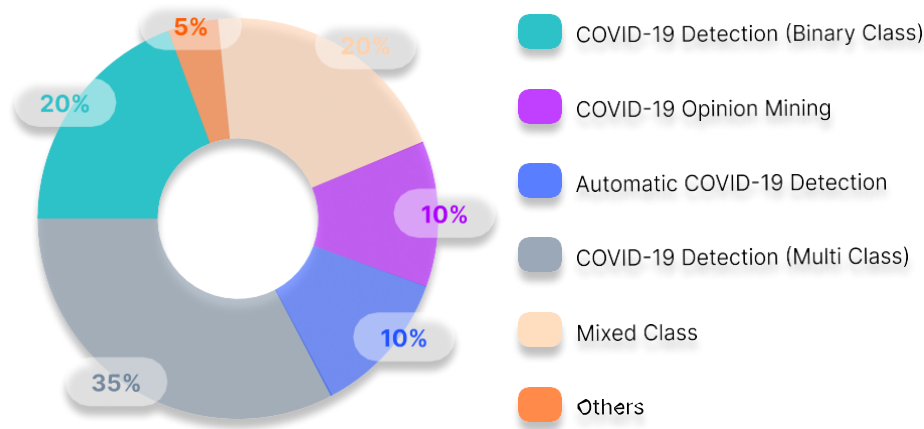

**Figure 5:** COVID-19 analysis amount based on application category.

Furthermore, open-access testing approaches have more far-reaching benefits when speeding up AI/ML/DL advances for the COVID-19 epidemic. Rapid dissemination of the COVID-19 pandemic has also exposed open source data as the latest major obstacle to the AI/ML-based COVID-19 battle Alimadadi and Cheng (2020), Shuja et al. (2021). Data balancing and curating are the most challenging issues, followed by data development. Ensure sufficient and large data Latif et al. (2020) for deep learning implementation (training, testing, and validation). Open-source data (text, speech, and image) and challenges are discussed in the following.

In the image-based analysis of COVID-19, data and information are not publicly accessible due to political concerns. It is highly recommended that the machine-related task be released accurately in a real-time application. Personal safety and cost with less labour should be ensured. As a result, more research is needed to create a connection between radio-graphic and PCR tests Alsabek et al. (2020). Most biomedical imaging research directions have data sets available. However, the scale of these data sets is limited for the implementation of deep learning techniques Ahsan et al. (2021). Deep learning algorithms need larger data sets, according to researchers, to provide more significant insights and precision in diagnosis Ardakani et al. (2020), Hemdan et al. (2020), Rana et al. (2023). Proper data augmentation and labelling are highly considered. Data sets and analysis focused on CT scans and X-rays are normal. At a higher cost, MRI offers high-resolution images with soft tissue contrast. A COVID-19 diagnostic and data set based on MRI is needed to equate its accuracy to CT scans and X-rays. Furthermore, it is important to examine how well the proposed AI/ML-based diagnosis would work and if it is possible to use it in clinical workflows while following rules and using fair evidence Bullock et al. (2020).

In speech-based analysis, the study on COVID-19 speech-based analysis of cough and breath rate effects is also in its early stages. Speech data should be noise-free and appropriately labelled to be successfully implemented in the deep learning model. Researchers have advocated for the compilation of voice evidence. But, because of the data used in ongoing experiments, only one open-access data set is available for speech-based COVID-19 diagnosis. Furthermore, current data collection sizes must be increased to improve classification process accuracy. Real-world data are highly selected for analysis, so a real-time system or method should be developed to generate data. Data preprocessing and feature extraction in speech data are complex Lella and Pja (2022), Sait et al. (2021), Pinkas et al. (2020).

Text-based data is used to predict the spread of COVID-19. Sometimes, it is also used for mining purposes from different clinical data on COVID-19. Some publicly available data can be collected, but sometimes, its analysis cannot validate a lack of imbalance. However, the use of NLP is suggested in these books. To solve these problems, COVID-19 datasets from social sites have been curated Alqurashi et al. (2020), Lopez et al. (2020). The spread of COVID-19 is analysed using deep learning, but correct and complete coverage data is needed. Blood test-based work is being done to analyse COVID-19, but the clinical data is rare and fully secure. Human attitudes and feelings during the pandemic, as well as feelings about lockdowns and other NPIs, have yet to be thoroughly studied. Another area of study is psychological distancing during the pandemic.

Furthermore, data set curation becomes needed to update performed, hence, social distances throughout COVID-19 lock-down initiatives. Another critical consideration for textual data is the timeliness of the study. Data pre-processing may affect performance. When social media data are collected, preprocessed, and annotated on a large scale, the interpretation and subsequent behaviour become rapidly obsolete Latif et al. (2020). Textual data for deep learning need to be large to train deep learning models Shuja et al. (2021).

## 2.3. Data pre-processiong tools for COVID-19 Analysis

Prior to its practical use, the data must be preprocessed. The purpose of data preprocessing is the transformation of raw data into a cleaned data set. Before running the method, the dataset is preprocessed to check for missing values, inconsistent data, and other irregularities. Data preprocessing is a crucial phase in the data mining process that includes transforming or removing data before it is utilised to ensure or improve performance. Table 5 is presented to show the list of preprocessing tools used in COVID-19 detection from image, text, and speech.

**Table 5**
Data pre-processing tools for COVID-19 analysis.

| Data Type | Pre-processing Tools | Purpose | References |
|---|---|---|---|
| Audio | Mel-frequency cepstral coefficients (MFCCs). | MFCCs are coefficients that collectively make up the Mel-frequency Cepstrum (MFC). | MFCC has primarily used terms in audio pre-processing Khriji et al. (2021). |
| | Fast Fourier Transform (FFT). | It breaks down a signal into its various spectral characteristics and, hence, offers frequency components. | FFT used in the preprocessing of breath sounds is used for the detection of COVID-19 Khriji et al. (2021). |
| | Discrete Cosine Transform (DCT). | A DCT is a method of representing a finite sequence of data points as a sum of cosine values of various frequencies. | DCT helps in recognition of COVID-19 disease from X-ray images with 2D curvelet transform Altan and Karasu (2020). |
| | The Short-time Fourier transform (STFT). | It is a Fourier-related transformation that can be used to identify the sine frequency as well as phase amplitude of local parts of signals as they change with time. | In the time of the COVID-19 pandemic, SFT was used to analyse the reliability of virtual tangible performance evaluations in veterans Ogawa et al. (2021). |
| Text | Lower casing. Removal of Punctuations. Removal of Stopwords. Removal of Frequent Words. Removal of rare words. Stemming. Lemmatisation. Removal of emojis. | Preprocessing your text implies converting it into a format that is predictable as well as analysable for your work. | pre-processing of text from Twitter and social media data helps to obtain good results in text-based COVID-19 analysis Paka et al. (2021), Nugraheni et al. (2020), Liang et al. (2021), Che et al. (2021), Islam et al. (2021b), Al-Waisy et al. (2023) Islam et al. (2023a), Contreras Hernández et al. (2023) Qorib et al. (2023). |
| Image | Pixel brightness transformations/ Brightness corrections, Geometric Transformations, Image Filtering and Segmentation, Fourier transform, Image Restoration, Normalisation, Standardisation, and Z-Component analysis. | To make images clearer and formatted so that machines can understand them, avoid irrelevant portions to get faster and more highly accurate results. | COVID-19 detection with fine pre-porocessed image using CT Deshpande and Schuller (2020), Subramanian et al. (2022) and X-ray image Khriji et al. (2021), Shorten et al. (2021), Jain et al. (2021), Sitaula and Hossain (2021), Ozyurt et al. (2021), Zebin and Rezvy (2021), Ahsan et al. (2021), Ardakani et al. (2020), Hemdan et al. (2020) Al-Waisy et al. (2023) Islam et al. (2023a), Khan et al. (2023). |

1. Preprocessing audio data can speed up the process more smoothly. Data cleaning, data aggregation, data reduction, and data transformation are the four stages of audio data preprocessing. The audio waveforms are processed to provide a topological map of sounds as a function of the time-lapse. The sample rate (n attribute values per second), amplitude (dB), and frequency are all metrics used to quantify the physical qualities of an audio signal (Hz). The most common and concise format of the audio signal is MFCC-dependent spectrogram transformation. There are a plethora of audio features available on the highest level of MFCC

Khriji et al. (2021). A variety of data points can be seen here. STFT Ogawa et al. (2021) is a compact sample size, but it's not really about compression; it's more about diverse ways of organising the same data on multiple axes or in a different space, but if you use MelSpectrogram, you'll most often reduce the size of the data point it into something relatively small, which is important when training a model. Another useful audio preprocessing method is the rapid Fourier transform (FFT) Khriji et al. (2021). The Fourier transform is generated using the FFT, which is a computationally efficient process. An FFT's key benefit is speed, which it achieves by reducing the number of computations required to evaluate a waveform. It breaks down a signal into its various spectral characteristics and hence offers frequency components. Transform of a Discrete Cosine (DCT) Altan and Karasu (2020). To generate the MFCCs, this reflects the Mel spectrum in the time domain. It has also been discovered that manipulating the MFCCs improves the model's accuracy. The speech model coupled the MFCC with the LSTM model, resulting in a high level of COVID-19 detection accuracy Khriji et al. (2021).

2. The actions taken to prepare pictures before they can be utilised in model training and inference are known as image preprocessing. This covers resizing, orienting, and colour corrections, among other things. Generally, it includes but is not limited to resizing, orienting, and colour corrections. There are some popular image pre-processing tools like Pixel brightness transformations/ Brightness corrections, Geometric Transformations, Image Filtering and Segmentation, Fourier transform, Image Restoration, Normalisation, Standardisation Z-component analysis, etc.

3. Simply put, preprocessing your text implies converting it into a format that is predictable and analyzable for your work. The most common text preprocessing tasks are: Lower casing, Removal of Punctuations, Removal of Stopwords, Removal of Frequent words, Removal of Rare words, Stemming, Lemmatization, and Removal of emojis.

## 3. Deep learning Methodologies used for COVID-19 Detection and Analysis

This section will discuss the most recent deep learning method for COVID-19 analysis and detection. We categorise the deep learning method for COVID-19 analysis and prediction into three categories: the deep learning model based on image, text, and speech data. This section discusses each model with details on figures, equations, analysis, and detection. CNN and RNN are basic deep learning models to analyse text, audio, and images. Some other types of deep learning models are Graph convolutional Networks, Generative Adversarial Networks (GAN), Attention-based mechanisms, capsule networks, CNN, and RNN. We will give internal details of each model, especially CNN. Before going into a deep description, we briefly describe the basic steps or prerequisites of COVID-19 analysis based on image, text, and speech.

### 3.1. Basic steps in Deep Learning Method for COVID-19 analysis from image

Deep learning-based image analysis for diagnosis and prognosis is very popular and established Talukder et al. (2022, 2023); Akter et al. (2024); Akhter and Hossain (2024). The general operation of a deep learning-based COVID-19 diagnosis and detection system is illustrated in the following Figure 6. Image-based COVID-19 analysis is the most popular and accurate analysis of COVID-19 Shah et al. (2021), Ardakani et al. (2020). In image-based analysis, a few basic steps are shown in Figure 6. Firstly, data are collected from open sources or crowd sources. Patients from the hospital community are considered participants during the data collection stage. Data can take various forms, but imaging techniques such as CT and X-ray samples can diagnose COVID-19.

COVID-19 analysis based on the image data in labelled form is fed to a different preprocessing method. There are some sub-steps in the data pre-processing steps like Normalisation, Standardisation, Zero Component Analysis, etc. Each image, along with CT images and X-ray images of COVID-19, has patient information about their heart. Data preprocessing steps remove an unwanted or irrelevant portion from images. Pre-processed data is sent to feature analysis. Feature analysis includes feature extraction, feature selection, and feature matching. After feature analysis, machine learning algorithms analyse the data and classify COVID-19 cases. There are sub-steps in the classification steps: data partition, model or algorithm selection, parameter tuning, and classification. The grouping phase divides the data for the experiment into preparation, validation, and testing sets. The cross-validation method is commonly used for data partitioning. The training data are used to create a specific model, which is then validated using validation data, and the output of the created model is tested using test data. The feature extraction and classification method is a key phase in the diagnosis based on deep learning of COVID-19. During this time, the deep learning methodology automatically extracts the attribute by repeating multiple operations, and finally, classification is performed based on the target or output class (Non-COVID-19 or COVID-19). Finally, the system is evaluated by metrics such as accuracy, recall, sensitivity, specificity, precision, and F1 score.

### 3.2. Basic steps in Deep Learning Method for COVID-19 analysis from speech

Identifying COVID-19 by analysing speech audio recordings containing coughs, breath sounds, or both remains a significant challenge due to the prevalent noise interference. The core obstacles in speech data analysis lie in distinguishing pertinent signals among the background noise inherent in coughing and breathing sounds. However, the scientific community has acknowledged the potential of sound as a valuable indicator of health for quite some time. For example, a built-in internal or external microphone receiver with recording is used in digital stethoscopes to detect sounds from the heart or lungs. Sound collection may be performed with mobile applications or recording-based medical instruments Lella and Pja (2022). These also refer to highly trained medical experts and experts to hear interpretations. These are recently and quickly being replaced by different image-based tools, such as various imaging techniques (e.g., MRI, radiography, sonography), with which it is easier to examine and interpret. Image-based data are precisely composed of actual data to be analysed. However, recent automatic audio interpretation and modelling tasks can diversify this trend and offer sound as an inexpensive and easily distributed alternative. In recent times, microphones have been used for sound processing on commodity devices such as smartphones and wearable devices Lella and PJA (2021).

In speech-based analysis, a few basic steps are shown in Figure 6. Firstly, data are collected from crowd sources or open sources. COVID-19 case analysis based on speech-based data with labelled form is fed to a different pre-processing method. Each data point has sound words or information on COVID-19 patients information. Data pre-processing steps remove unwanted or irrelevant portions. MFCC and FFT have been used terms mainly in audio pre-processing Khriji et al. (2021). The preprocessed data are sent to a machine learning algorithm to analyse the data and classify COVID-19 cases. In the final step, COVID-19 positive and negative cases of COVID-19 are found with accuracy, sensitivity, precession, etc.

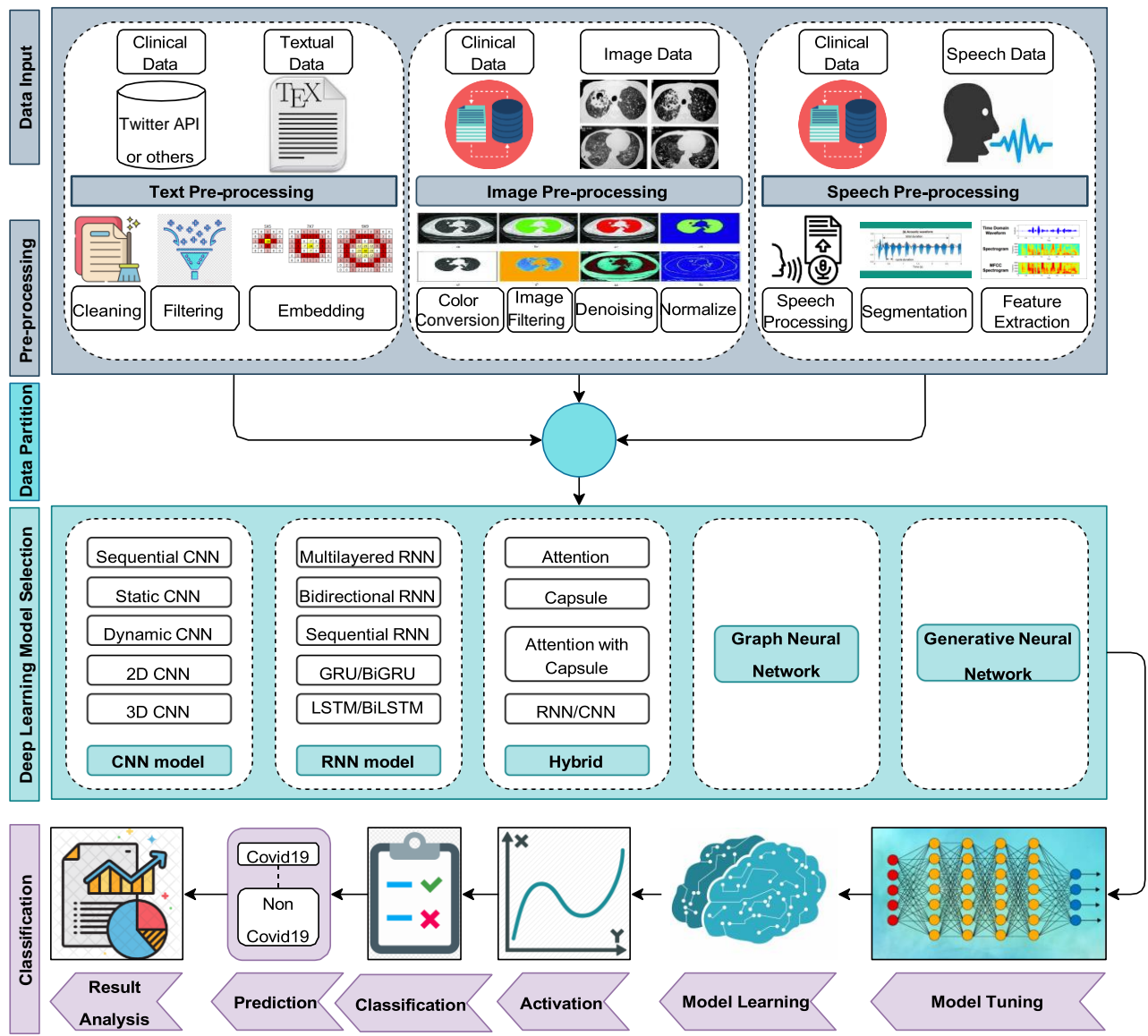

**Figure 6:** Overall diagram of all deep learning methods to analyse COVID-19 from text, image, and speech.

There are a lot of deep learning-based speech recognition systems. Most approaches use conventional deep learning models or hybrid deep learning models to recognise speech towards end-to-end speech recognition with RNN Graves and Jaitly (2014). This method provides a speech recognition method that directly transcribes speech or audio data with text without requiring an intermediate phonetic presentation. The system combines the deep bidirectional LSTM of recurrent neural network architecture. A method that is end-to-end deep learning and RNN strategy can be utilised to recognise either English or Mandarin Chinese speech, even though the two languages are substantially different. End-to-end learning allows us to manage a wide range of speech, including loud surroundings, accents, and multiple languages, because it transforms entire pipelines of hand-engineered elements with neural networks Amodei et al. (2016).

On the other hand, for automatic voice recognition, an acoustic-based work offers a new end-to-end neural acoustic model. The model is made up of several blocks that are connected by residual connections. Each block comprises one or more modules with separable convolutional layers in 1D time channels, batch normalisation, and ReLU layers. CTC loss is used to train it Zhang et al. (2017). Another new hybrid method is CNN-DNN-HMM combined time convolution for speech recognition in the field Yoshioka et al. (2015).

### 3.3. Basic steps in Deep Learning Method for COVID-19 analysis from text

While we have explored deep learning techniques in the preceding section, it's worth noting that methods such as LSTM, GRU, and CNN are widely employed in text-based COVID-19 analysis for detection and prediction. This section outlines the fundamental structure and procedural steps involved in using deep learning for text-based COVID-19 analysis. It is important to recognise the multitude of deep learning methodologies available for COVID-19 analysis. In text-based analysis, a few basic steps are shown in Figure 6. COVID-19 case analysis based on textual data is fed to the different pre-processing methods. Each data point has a keyword or information about COVID-19 patient information. Data preprocessing steps in the cleaning of unwanted or irrelevant information. After preprocessing, the embedding layer converts each text term to a word vector. The preprocessed data are sent to a machine learning algorithm to analyse and classify COVID-19 cases. The most widely used and popular text-based deep learning method for COVID-19 analysis is LSTM. Time-aware based LSTM for COVID-19 analysis and prediction gives 98% accuracy Li et al. (2021). Detecting fake news for the COVID-19 BiLSTM-based method shows 95% accuracy Paka et al. (2021). On the other hand, the RNN model with attention and capsule network mechanism provides successful results Islam et al. (2020a). A GRU-based method obtained 96% accuracy for classifying public behaviours in the era of COVID-19 Raamkumar et al. (2020).

Figure 6 presents the overall diagram of all deep learning methods to analyse COVID-19 from text, image, and speech. At the initial stage of the figure, different data types are passed to the input and then pre-processed based on the data type. After the pre-processing level, the deep model is executed. Here, we see that the deep learning model is classified into four parts: RNN, CNN, Hybrid, and others. If multiple layers of CNN or RNN are in a series pattern, then it is called sequential, which can also be used stacked.

### 3.4. Convolutional Neural Network (CNN) Method for COVID-19 analysis

CNN, or the Convolutional Neural Network, is frequently used in deep learning for image analysis, employing convolutional layers to extract features from visual data. CNN is also used in text and speech classification. Lecun first proposed the CNN structure in 1998 LeCun et al. (1998). A convolution stratum converts small data pieces into standard high-level vectors. Figure 7 shows the basic setup and related

CNN layout layers. CNN is sufficiently stable and gives a more extensive scope and input. The inputs of CNNs are entirely mapped to the output layers. CNN also has layers for convolutional and pooling layers that are essential for CNN performance. For each feature, these layers require filtering maps. The feature maps represent the source information and the filters used in the backpropagation task through convolution and bundling layers. The actual superiority of CNN over its counterparts is that, without human supervision, it dynamically determines the related facets. For example, given appropriate images of cats and dogs, the key features for each class would be learned by itself. CNN does not encode the location and orientation of the object in its projections. They completely ignore all their inner awareness and orientation and pass all information to the same neurones that cannot handle it.

This approach was used to analyse COVID-19 from image Islam et al. (2020a), Sohrabi et al. (2020), Luz et al. (2021), from text Nugraheni et al. (2020) and speech data Sait et al. (2021), Lella and Pja (2022), Brown et al. (2020). Figure 7 gives an overall picture of COVID-19 analysis using all CNN-based models from text, image, and speech data. In this figure, three data types are inputted separately through the preprocessing steps and are forwarded to the CNN-based transfer learning model. The result of model execution is forwarded to the pooling, flattened, fully connected, and SoftMax layer sequentially to predict the covid or non-covid class. All CNN models (transfer-based) are presented, and their internal structure is highlighted in Figure 7.

In this section, each model using CNN will be discussed with an internal description of the figure, recent works, and applications.

- LeNet: LeNet CNN is a pioneering architecture for image, text and speech recognition, notably excelling in hand-written digit recognition tasks. Yann LeCun proposed the LeNet CNN structure in 1989 LeCun et al. (1989). LeNet is a basic convolutional neural network commonly referred to as LeNet-5. CNN Convolutional neural networks (CNN) are feed-forward neural networks with artificial neurones that can interact with a portion of the surrounding cells in the coverage area and handle large-scale image analysis. The LeNet-5 denotes CNN's emergence and identifies its main components. However, it was not widely used because of a shortage of hardware, mainly GPUs. A recent approach to analysing COVID-19 with the LeNet model obtained an accuracy of 56.06% Islam and Matin (2020). The basic structure of LeNet is shown in Figure 7 to analyse COVID-19 from text, image, and speech.

- AlexNet: is a groundbreaking convolutional neural network pivotal in advancing image classification, notably winning the 2012 ImageNet Large Scale Visual Recognition Challenge. AlexNet architecture was created by Alex Krizhevsky with his team Krizhevsky et al. (2012). AlexNet had eight layers: the first five were convolution layers, followed by max-pooling layers in some cases, and the final three were fully linked layers, as shown in Figure

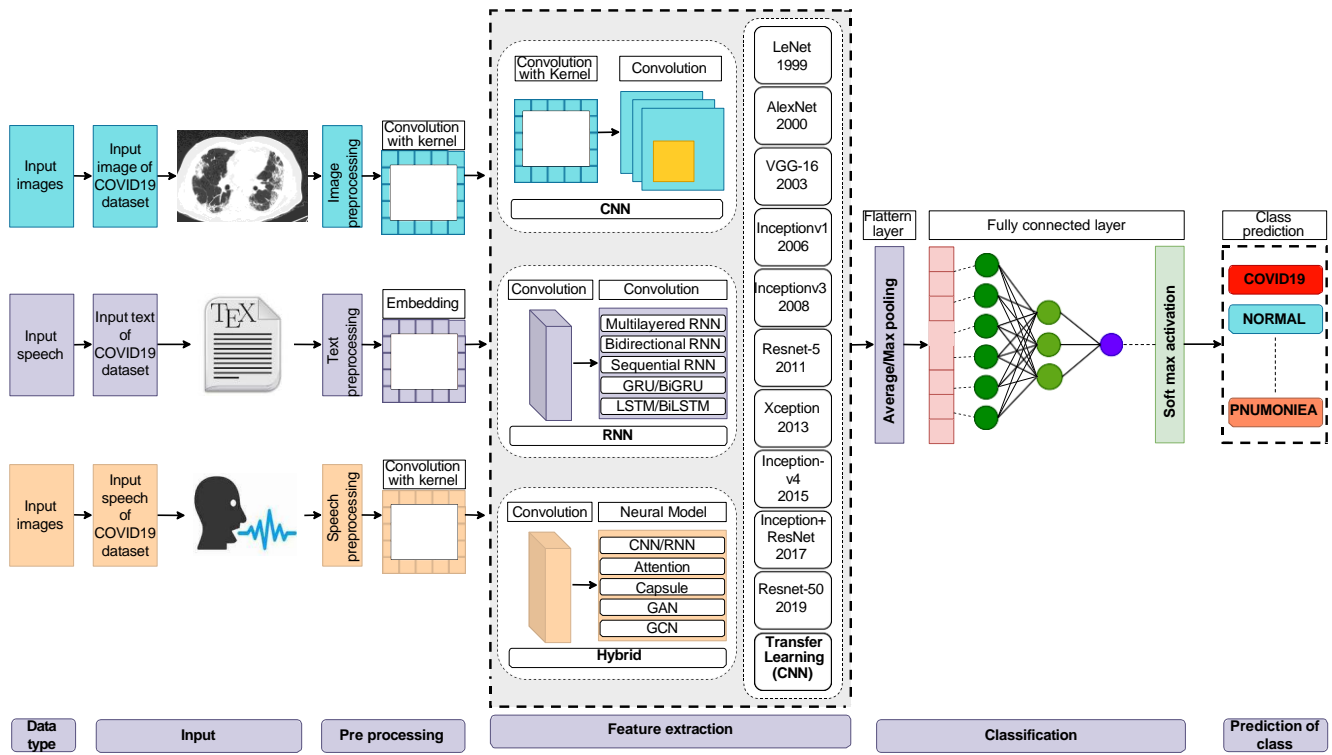

**Figure 7:** Basic model of CNN model, RNN model, Hybrid and Transfer learning model for COVID-19 analysis from image, text, and speech.

7. It used the non-saturating ReLU activation function, which outperformed tanh and sigmoid in training accuracy. Image-based COVID-19 analysis is done with AlexNet Ardakani et al. (2020). The basic structure of AlexNet is shown in Figure 7 to analyse COVID-19 from text, image, and speech.

- **VGG:** Oxford Robotics Institute created the Visual Geometry Group Network (VGG) based on the fully convolutional network architecture Simonyan and Zisserman (2014). In 2014, the Wide Visual Recognition Challenge was also discussed (ILSRVC2014). The VGGNet used smaller filters of 3 to 3 compared to the AlexNet 11 to 11 filters to increase the image extraction capabilities. VGG has two versions: one is VGG-16, and the other one is VGG-19. The performance of the VGG model to detect COVID-19 is good. Recently, COVID-19 has been analysed by ResNet Ardakani et al. (2020). The basic structure of VGG-16 is shown in Figure 7 to analyse COVID-19 from text, image, and speech.

- **DenseNet:** The "Dense Convolutional Net" has many persuasive effects: it lightens the challenge, strengthens the dissemination of features, encourages the reuse of features, and significantly reduces the number of parameters Huang et al. (2017). One of the most recent developments in neural networks for visual object detection is DenseNet. DenseNet is quite like ResNet, although there are a few key distinctions. DenseNet concatenates the prior layer (identity) with the upcoming layer, whereas ResNet utilises an additive approach that combines the previous layer (identity) with the next layer. Using DenseNet, COVID-19 detection is done with successful results Horry et al. (2020). The basic structure of VGG-16 is shown in Figure 7 to analyse COVID-19 from text, image, and speech.

- **InceptionNet:** In a CNN, an Inception Module is an image model block whose goal is to evaluate approximately an ideal local sparse figure. Simply said, it lets us employ many sorts of filter sizes in every image block rather than being limited to a singular filter size that we then concatenate and pass onto the next layer. Inception network, or GoogleNet network of 22-layer networks, was successfully established in 2014 and got a top five precision of 93.3% Szegedy et al. (2016a). Subsequent versions are called Inception VN, where N is the version number. Using Inception, COVID-19 detection is done with successful results Horry et al. (2020). The basic structure of VGG-16 is shown in Figure 7 to analyse COVID-19 from text, image, and speech.

- **ResNet:** By using skip connections that jump over many network layers to achieve good convergence behaviour, ResNet, short for Residual Networks, is a popular neural network that serves as the foundation for many computers' vision and speech recognition tasks. In 2015, this model was the champion of the ImageNet challenge. ResNet was very outperformed because it successfully trained extraordinarily deep neural networks with 150+ layers. Szegedy et al. (2016b) introduced residual network (ResNet) models. ResNet's upgraded variant is ResNet-V2. The ResNet is close to VGGNet and about eight times more profound Too et al. (2019). Recently, COVID-19 has been analysed by ResNet Ardakani et al. (2020) Şengür et al. 2021 with almost 95% accuracy. Figure 7 shows the basic structure of the ResNet model for COVID-19 analysis from image, text, and speech. The basic structure of ResNet is shown in Figure 7 to analyse COVID-19 from text, image, and speech. Some recent methods of COVID-19 are performing better than other methods like ResNet Al-Waisy et al. (2023), ResNetv3 Islam et al. (2023a),

- Inception-ResNet-V2: It is a combinational hybrid model with InceptionNet and ResNet. It is a 164-layer, profound neural network incorporating the architecture of Inception and residual relations. The variant of Inception V3 Szegedy et al. (2017) is Inception-ResNet-V2. Start-up ResNet-V2 has been educated in the ImageNet database on more than a million files. Using the Inception and ReNet model together, COVID-19 detection is done with good results Horry et al. (2020). The basic model is shown in Figure 7.

- XceptionNet: The feature extraction base of the Xception framework is made up of 36 convolutional neural networks. The Xception model structure is a linear pile of depth-wise convolutional layers with feedback blocks for easy definition and modification of deep-network architecture Chollet (2017). The Xception is an improvement in the design of the Genesis, which substitutes standard initial modules with distinctive depth convolutions. Using Xception, the COVID-19 detection is done with quite a successful result Horry et al. (2020). The basic structure of XceptionNet is shown in Figure 7 to analyse COVID-19 from text, image, and speech.

- MobileNet: MobileNet is a simplified architecture that constructs simple deep neural networks using depth-wise separable CNN layers and provides a good model for mobile and embedded vision-based applications. By decreasing the quantity and adding the reversed residual with the linear bottleneck blocks, MobileNetsachieves accomplishes this crucial advantage of significantly reducing memory usage. In addition, Mobile Net V2 is commonly used in many common deeper learning frameworks as a pre-trained implementation [99]. A new application of MobileNet to detect COVID-19 Ardakani et al. (2020), the basic operational sequence of MobileNet is shown in Figure 7 to analyse COVID-19 from text, image, and speech.

- Google Net: GoogleNet is a 22-layer deep convolutional layer provided by Google researchers as a variety of the Inception Network, a Deep Convolutional Neural Network. The GoogleNet convolutional neural network is based on the Inception design and is a convolutional neural network; it is shown in Figure 7. It employs Inception modules, enabling the network to select from various convolutional filter sizes in each block. At this time, GoogleNet is working well to detect COVID-19. The basic structure of GoogleNet is shown in Figure 7 to analyse COVID-19 from text, image, and speech.

□ Convolutional Neural Network (CNN) Method for COVID-19 analysis from the image:

The CNN-based method is very popular and convenient in image classification. In recent times, CNN-based methods have been widely used in COVID-19 detection and classification from images. A most recent approach in COVID-19 detection with CNN classifier by Jain et al. (2021). This approach achieves an average accuracy of 97%, but this method works on small data and has an overfitting problem. CNN's leNet-based classifier was used to classify small CT image data with 56.06% Islam and Matin (2020). However, this LeNet is not good for large data sets. Zebin et al. analysed 822 X-ray images to detect COVID-19 and obtained an accuracy of 90%, 94.3%, and 96.8% with VGG16, ResNet50, and EfficientNetB models, respectively Zebin and Rezvy (2021). However, this method is complex to operate. Another method with ResNet and SVM got 94.7% accuracy, but this method works with a small dataset. Multifunction-based deep learning method with ResNet50 to analyse clinical data for screening COVID-19 with Accuracy of 76%, Sensitivity of 81.1%, Specificity of 61.5%, and AUC of 81.9 %. COVID-19 detection was done with extensive statistical type analysis using the ResNet52 model with an accuracy of 94.98. However, this method extracts only the attentional region rather than the lesion segmentation. Another study that used almost all CNN-based classifiers showed their performance and suggested the ResNet model as the best model Ardakani et al. (2020). Coronet by Sohrabi et al. (2020) is another recent method to analyse COVID-19 with an accuracy of 89.5%, a sensitivity of 100%, a precision of 97%, and an F1-Score of 98%.

□ Convolutional Neural Network (CNN) Method for COVID-19 analysis from text:

CNN-based model is widely used in image and text classification. A recent method to classify the aggravation status of COVID-19 events from short text using CNN. This method classified 16833 tweets and got an accuracy of 94.99% Nugraheni et al. (2020). Another CNN method, along with the RNN method, is to analyse misinformation on Twitter. This method handled 3668 texts and got precession 71.84%, recall 70.98%, and an f1 score of 71. 40 % Islam et al. (2021b). Data annotation should be improved for this method.

□Convolutional Neural Network (CNN) Method for COVID-19 analysis from speech:

CNN is a common method for speech classification and recognition. The CNN inception model used 4558 speeches to detect COVID-19 from breathing sounds. This method got 80% accuracy in Fig. 7. The performance of this method for speech analysis is limited to the pre-processing of speech. The data set is very small. VGG model to detect COVID-19 from cough sound 660 sounds. This method got 80% accuracy, 72% precession and a recall of 69%. The most powerful method in speech analysis is ResNet, a new method with ResNet to detect COVID-19 from 5210 recordings. This method got a sensitivity of 98. 5%, a specificity 94% Laguarta et al. (2020b). This is a complex model and takes slightly more training time.

## 3.5. Recurrent Neural Network (RNN) Method for COVID-19 analysis

RNN, or Recurrent Neural Network, processes sequential data with connections forming loops to retain the memory of past input. RNN can handle long-ranged features in data analysis. The basic structure of the RNN-based COVID-19 analysis is shown in Figure 7. A recurrent neural network (RNN) is essentially a form of artificial neural network (ANN) that creates a direct graph over a series of time connections between nodes. RNNs may use their hidden representation (memory) to manage arbitrary input length ranges developed by feedforward neural networks (FNN)Tealab (2018). The RNN refers to the finite and infinite impulses of two different groups of networks with a similar general structure. The additional retained condition will include all limited impulses and infinite impulse repeated networks. Often, any other network or chart may replace storage. These have delays as well as feedback loops. This controlled state is called a gated state or gated memory and is used in all long-term memory networks (LSTMs) and recurrent gated units. It is also called the Neural Network of Feedback (FNN).

### 3.5.1. Long Short-Term Memory (LSTM) Method for COVID-19 analysis

Long Short-Term Memory, or LSTM, is a special kind of recurrent neural network that was made to fix the problem of vanishing gradients. This makes it easier to learn about long-term dependencies. LSTM is usually an RNN extension that requires long-term storage of the inputs. Compared to the basic internal memory of RNN, LSTM has advanced memory. You can read, write, and remove the memory material. Then, it tackles the disadvantage of RNN, which has been affected by its disappearance. LSTM determines which information is to be remembered and forgotten. LSTM can hold the memory. There are three doors: input, forget, and exit doors.

### 3.5.2. Gated Controlling Units (GRU) Method for COVID-19 analysis

GRU, or Gated Recurrent Unit, is a simplified yet powerful recurrent neural network architecture for processing sequential data. Kyunghyun developed Gated Controlling Units (GRU) as a Current Network (CNN) Cho et al. (2014) cross-recourse feature. LSTM and GRU are very much in sequence. The specifications of GRU are limited to LSTM. More than LSTM, GRU works. GRU uses a high-efficiency gate. GRU includes two resets and upgraded doors. Each reset gate regulates how the input of new information is combined with the previous information, and the update gate defines the preceding memory. No context conditions (ct) like LSTM have been developed for GRUs. Updating and resetting the gate allows the gate to forget and be attached to the secret layer of the past. The LSTM reset gate is then effectively divided into the reset and update elements in a GRU.

□Recurrent Neural Network (RNN) Method for COVID-19 analysis of speech:

Sequential LSTM with a convolutional network is doing an excellent job of detecting COVID-19 from images. COVID-19 Case Prediction Using LSTM and RNN: Lethality and Tests in GCC Countries and India The LSTM model with numerous hidden layers has a high level of accuracy and a small prediction error Razia et al. (2021). Another RNN-based method with CNN got 99.4% accuracy from 4575 images to detect COVID-19 Islam et al. (2020b).

□ Recurrent Neural Network (RNN) Method for COVID-19 analysis from text

Clinical data is analysed with time-sensitive LSTM to predict COVID-19 disease progression and patients from 3120 data samples. This method got the highest accuracy of 98 for a specific time. This method is complex because it depends on multiple features and makes false predictions at a certain time Li et al. (2021). Good work with the LSTM model for COVID-19 Fake News Detection with 10700 data obtaining 98.59% f1 score Liang et al. (2021). This method is very simple and needs to be improved in modelling analysis. A new method with LSTM to analyse the emotions and symptoms of the Arabic people of 300000 tweets data at the time of the COVID-19 pandemic Al-Laith and Alenezi (2021). The accuracy of this method is 82.60%, so it should be improved. Cross-cultural polarity and emotion detection are other methods to analyse COVID-19 tweets Imran et al. (2020). This method got 82.40% accuracy, and its accuracy and adaptability should be improved. A GRU-based method to classify public behaviour towards physical distancing Raamkumar et al. (2020). This method classifies 16752 comments and gets 96% specificity. The stability of this method should be improved.

□Recurrent Neural Network (RNN) Method for COVID-19 analysis of speech

Generally, this approach works well to analyse COVID-19 Islam et al. (2020a). To achieve higher performance, capsule networks and attention approaches can sometimes be added with RNN. The Capsule-based approach reduces the training time and was recently used in COVID-19 analysis Afshar et al. (2020). There are also other methods, except for CNN and RNN; all related models will be discussed in this section based on the data. Another method with the LSTM model was to classify 320 patient audio samples and obtain accuracy in Cough Sound 97%, Breath Sound 98.2%, and Voice 84.4% Hassan et al. (2020). Smartphone Recordings for COVID-19 Cough Classification with LSTM got 94% accuracy with 1079 samples Pahar et al. (2021).

## 3.6. Attention Method for COVID-19 analysis

Attention processes became more prominent in NLP after an important paper on machine translation Bahdanau et al. (2014). It acts as a single hidden layer on a neural net to grasp the attention process clearly. The attention module's goal is to determine the importance of each hidden layer and generate a weighted total of all input data. In recent times, during the era of COVID-19, cough and breath-based speech data have been used to analyse and predict COVID-19. Here, a new method or speech-based COVID-19 analysis using attention-based GRU got 87% recall Pinkas et al. (2020).

□ *Attention method for COVID-19 analysis from image*
Attention-based deep learning models can handle features more effectively than other methods. Attention-based VGG-16 classifier to detect COVID-19 from 8571 chest X-ray images Sitaula and Hossain (2021), This method got an accuracy of 87. 49%. Data Augmentation should be improved with this method. Dense-based GAN and multi-layer attention model with a good segmentation method for COVID-19 CT images with higher precession of 94.6% Zhang et al. (2021a).

□ *Attention method for COVID-19 analysis from text*
This method works on minimal data, and the model structure should be improved and straightforward. Attention-based BiLSTM to find COVID-19 fake news from 16000 tweets. This method got 95.40% accuracy but cannot handle retweets sufficiently Paka et al. (2021). Another method to classify COVID-19 Chinese 25392 text sentiments with attention-based BiLSTM and CRF Zhang et al. (2020). This method got a Precession of 85.49 %, recall of 83.14%, and F1 score of 84.30%. This model's performance depends on features, so it should be more adaptable.

□ *Attention method for COVID-19 analysis from speech*
A method for abnormal patient classification with real-world audio using attention-based BiGRU Wang et al. (2020b). This method got 94.5% accuracy. It works on very small data, and the internal structure of this method is unclear. Another method with attention-based GRU was to analyse 92 audio samples to detect COVID-19 and obtain 87% accuracy. The accuracy of this model is not good and the correlation of characteristics can lead to miss some important features; the data sample should be extended Pinkas et al. (2020). BiGRU with an attention-based model to detect COVID-19 from 4217 audio samples Dastider et al. (2021). This method got 83.7% accuracy. Data samples should be extended, and precision should be improved by improving model learning.

### 3.7. Capsule Network Method for COVID-19 analysis

Capsule networks introduce hierarchical relationships between parts of objects, potentially advancing image recognition beyond traditional convolutional neural networks. The capsule and its primary capsule layer provide a profound convergence for capturing and maintaining a dense hierarchy parallel to the different dimensions of the kernel Mobiny et al. (2020). The output is sent to the secondary layer of the capsule network. The secondary capsule layer uses its dynamic routing strategy to assess the essential functions and performance of each capsule. The performance of the secondary capsule is transmitted into the global pool layer to optimise the activations achieved. The last layer is a fully connected layer, with the result class options being activated simply in sigmoid form.

Sometimes, a capsule network approach can be added with RNN to get higher performance. The Capsule-based approach reduces the training time and was recently used in COVID-19 analysis Afshar et al. (2020).

□ *Capsule Network Method for COVID-19 Analysis from Image*
A capsule network can reduce training time. Diagnosis of COVID-19 from X-ray images with capsule network from 13800 images of COVID-19 data with 95.7% accuracy Afshar et al. (2020). A new Radiologist COVID-19 detection of CT scan image with complete capsule networks with an accuracy of 87.6% Mobiny et al. (2020). This model works only on a small domain with 746 images.

□*Capsule Network Method for COVID-19 Analysis from Text*
COVID-19 outbreak analysis by an ensemble pre-trained deep learning model for detecting informative tweets Malla and Alphonse (2021). A deep learning model for text classification using a capsule with a simple routing strategy effectively reduces the computational Kim et al. (2020) complexity of dynamic routing. Additionally, proposed static routing, an alternative to dynamic routing, results in higher accuracies with less computation. Sentimental Analysis using Capsule Network with Gravitational Search Algorithm. In COVID-19 text data mining, the Capsule network works excessively for sentiment analysis compared to other models Prabha and Rathipriya (2020).

□ *Capsule Network Method for COVID-19 analysis from speech:*
We studied a lot, but did not do well. Enough research on speech-based COVID-19 analysis using the capsule network Multitask learning can be incorporated into capsules, which can help a model perform better when the task is challenging. The fundamental capsule network was given a regularisation to add structure to its output: it learns to recognise the speaker of the speech by forcing the needed details into the capsule vectors Poncelet et al. (2020). Some ideas for handling speech with capsule networks came from a review paper Shuja et al. (2021).

### 3.8. Generative Adversarial Network (GAN) Method for COVID-19 analysis

GAN, or Generative Adversarial Network, involves two neural networks competing to generate realistic synthetic data, commonly used for tasks like image generation and data augmentation. In 2014, Ian Goodfellow, with his colleagues, created the generative adversarial network (GAN) Goodfellow et al. (2014), a family of machine learning frameworks. GANs (Generative Adversarial Networks) have recently been used to create images, videos, and speech. GANs are algorithmic architectures that combine two DNN architectures to create new simulated data instances that can be applied to real data. GAN, or Generative Adversarial Network, is a neural network made up of two competing deep learning models: the generator and the discriminator. Their goal is to generate data sets similar to some of the training set's data sets. Figure 8 shows the basic GAN model for the analysis of COVID-19 from image, text, and speech.

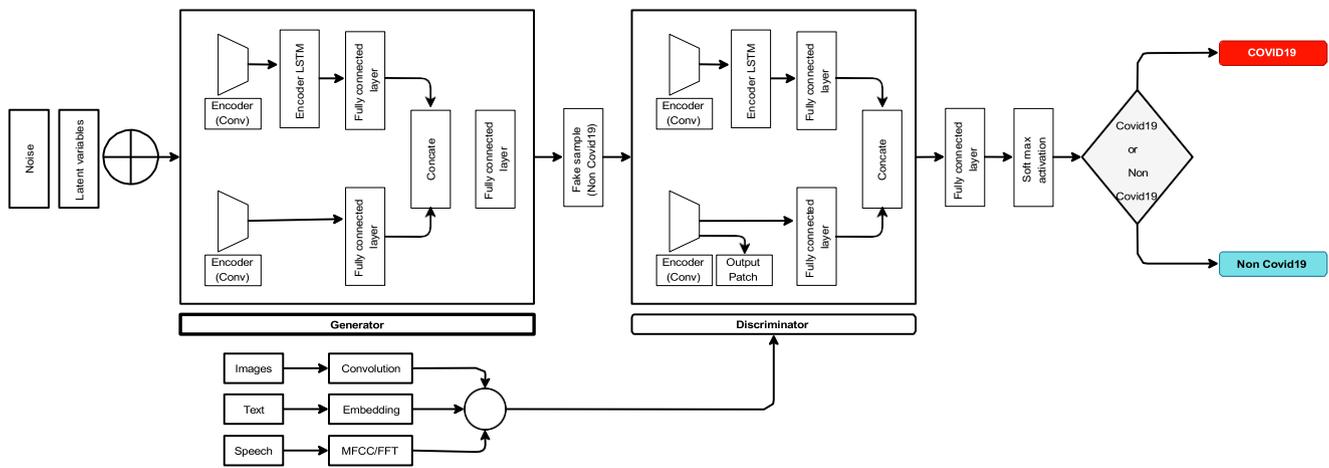

**Figure 8:** GAN model for COVID-19 analysis from image, text, and speech.

☐ *Generative Adversarial Network (GAN) Method for COVID-19 analysis from image:*

The author Waheed presented GAN architecture for a small COVID-19 CXR dataset of 192 images, and their structure achieved levels of accuracy of 85.00% and 95.00%, respectively, pre and post-GAN augmentation Elzeki et al. (2021). A new computer-aided deep learning model for GAN classification shows good performance in three different datasets Bar-El et al. (2021). The most recent method for automatic screening for COVID-19 uses an Optimised Generative Adversarial Network Goel et al. (2021).

☐ *Generative Adversarial Network (GAN) Method for COVID-19 analysis from text:*

From random noise, GANs can create accurate information. For time series prediction, this unique strategy that integrates data-corrected optimisation with GAN and BiL-STM offers promising results Quilodrán-Casas et al. (2022). An improved Cycle GAN with applied area to COVID-19 classification Bar-El et al. (2021). A very nice use of GAN has been made in this method to analyse COVID-19. A COVID-19 Hate Speech Twitter Archive (CHSTA) was used to scrape, ingest, and aggregate tweets, and then hate speech was assessed using the DCAP method inspired by the Generative Adversarial Network (GAN). From tweets in English about the COVID-19 pandemic, our approach can detect the occurrence and changing dynamics of hate speech Fan et al. (2020).

☐ *Generative Adversarial Network (GAN) Method for COVID-19 analysis from speech*

GAN model performed better than all other methods in COVID-19 analysis Jamshidi et al. (2020). COVID-19 infection has been shown to cause harm to human airway epithelial cells. It has also been established that the impact of human respiratory secretions on the human airway and the results of scanning electron microscopy and genetic analysis of the culture supernatant can be used to visualise and detect a novel human Coronavirus Zhu et al. (2020).

### 3.9. Graph Convolutional Network (GCN) Method for COVID-19 analysis

A graph neural network (GNN) is a type of neural network that uses graph data structures to analyse data Scarselli et al. (2008). Graph neural networks (GNN) seem to be neural models that use message transmission between graph nodes to capture graph dependency. The GCNN model shows that convolutional features are passed to graph convolution. After the graph operation, important features are selected with the normalisation of nodes. Figure 9 shows the basic GCN model for COVID-19 analysis of images, text, and speech.

☐ *Graph Convolutional Network (GCN) method for COVID-19 analysis from image*

Graph Convolutional Neural Network-based work with 296 CT images to detect COVID-19 Yu et al. (2021). This method performs higher with feature extraction and got 96.62% accuracy, but its domain is very small. In this work, the individual feature vectors and the relational features of both GCN and CNN were fused using deep feature fusion (DFF) Wang et al. (2021). This method works with a higher accuracy of 97.71%. This method can learn relation-aware representations.

☐ *Graph Convolutional Network (GCN) Method for COVID-19 Analysis from Text*

A GCNN method uses a spatial-temporal graph neural network to investigate the prediction of COVID-19 Kapoor et al. (2020). This method got RMSLE 0.0109 error. The Chinese text grammar was coupled with a graph convolutional neural network in a social network sentiment classification approach to get 4.45% margin accuracy [108]. Another new method with Graph convolutional neural network on 100000 entities COVID-19 analysis Che et al. (2021). This method uses a knowledge-based approach and has 95.40% accuracy. The performance of this model is limited to data labelling.

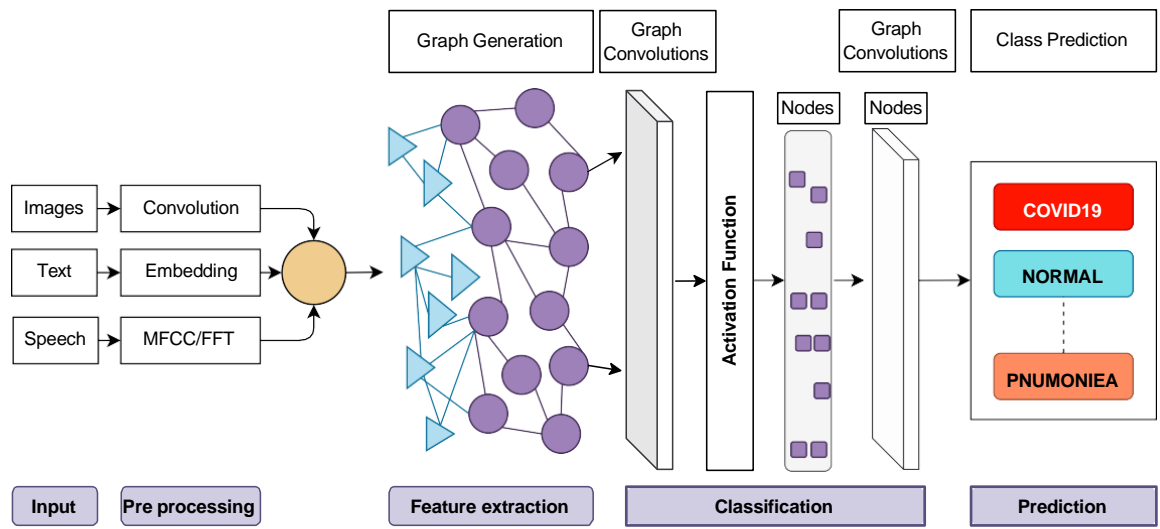

**Figure 9:** GCN model for COVID-19 analysis from image, text, and speech.

□*Graph Convolutional Network (GCN) Method for COVID-19 Analysis of Speech*

The graph-convolutional neural network (GCN) performs well in disease-based speech recognition and classification. Recent work is longitudinal Speech Biomarkers to Automated Alzheimer's Detection Laguarta et al. (2020a). GCN model is not popularly used in COVID-19 analysis with speech data. However, this method works well with images and text-based data. Three-dimensional deep learning model to diagnose COVID-19 Pneumonia Based on Graph Convolutional Network Liang et al. (2021).

## 3.10. Hybrid deep learning method for COVID-19 analysis

Hybrid deep learning methods fuse diverse architectures and techniques to improve performance in complex tasks. In fact, a hybrid deep learning-based method is developed using CNN, RNN, and other methods, as shown in Figure 7. Here, we present a short outline of the recently developed deep learning-based hybrid model to analyse COVID-19. Some of the current state-of-the-art image classifiers needed to complete the clinic intent are described in this section. This section shows a recently developed deep learning method for analysing COVID-19 from image, speech, and text data.

□ *Hybrid deep learning method for COVID-19 analysis from the image*

COVID-19 detection from chest 424 X-ray images using a novel Deep GRU-CNN model with a precision of 96.00% Shah et al. (2021). For the diagnosis and classification of COVID-19, a deep residual network-based type attention layer with bidirectional LSTM was developed. It is a new ResNet-based Class Attention Layer with a Bidirectional LSTM called RCAL-BiLSTM that achieves an accuracy of 94.88% for COVID-19 Diagnosis Pustokhin et al. (2020). Another hybrid deep learning method with 40000 images to detect COVID-19 Islam et al. (2020a). This method achieved 99.4% accuracy. A Deep Neural network model with ANN to COVID-19 with feature extraction works on 20000 images and achieves 95.84% accuracy Ozyurt et al. (2021). CNN-LSTM-based method to classify 4575 images to detect COVID-19 with 99.2% accuracy Islam et al. (2020b).

□ *Hybrid deep learning method for COVID-19 analysis from text*

CNN-LSTM method to forecast COVID-19 infected people with 90.63% accuracy Parashar et al. (2020). The description of the data set is not clear. Another method with CNN and RNN models Islam et al. (2021b). This method classifies 3668 images to detect COVID-19 with a precision of 71.84%. Data annotation of this method needs to be improved. Text embedding section and pre-processing handle the quality of performances.

□ *Hybrid deep learning method for COVID-19 analysis from speech*

Most conventional CNN and LSTM on 90 audio and video samples predict COVID-19 with 67% Dastider et al. (2021). The performance of this model depends on the convex probe and pre-processing. An attention GRU with generated embeddings from the audio inputs to detect COVID-19 Pinkas et al. (2020) with 87% accuracy. Speech recognition using CNN and BiLSTM Passricha and Aggarwal (2020). An encoder and decoder with a multilayer perception model classify 1044 images with 86% accuracy Amyar et al. (2020). This method segments the regions of interest simultaneously. Another hybrid method Saood and Hatem (2021) for detecting COVID-19 from a lung CT picture using segmentation and deep learning, U-Net, as well as SegNet. This method got 95% accuracy but has some time complexity and worked on small data.

## 4. Comparative and analytical Result analysis

In this section, we discuss the comparative result. This section gives a complete analysis of the results in two parts. The first is a comparative result analysis of recent methods by reviewing, and the second is an analytical result analysis by our manual implementation of methods.

Some performance metrics are used to evaluate the performance of the deep learning model. Here, we present the most popular evaluation metrics. In the calculation, P indicates Positive, T indicates True, F indicates False, and N indicates Negative. In the loss function, y is the predicted value $y_i$. Their calculations are computed as follows in equations (1) to (5).

$$Accuracy : \frac{(TP + TN)}{TP + TN + FP + FN} \tag{1}$$

$$Precision : \frac{(TP)}{TP + FP} \tag{2}$$

$$Recall : \frac{(TP)}{TP + FN} \tag{3}$$

$$F1 : 2 * \frac{(Precession * Recall)}{Precession + Recall} \tag{4}$$

$$Loss(y) : y_i log(\bar{y}_i) + (1 - y_i) + log(1 - \bar{y}_i) \tag{5}$$

### 4.1. Comparative result analysis of recent methods

We have added an overall comparative result analysis, which is one of the important parts of this article. We critically analysed the most relevant papers in the comparative analysis to clearly show their findings and research gaps.

We have also added information about the method, data, validation method, etc. In the main comparative Table 6, CV means Cross-Validation, A means Accuracy, S means Specificity, P means precision, R means Recall, F for F1 score, T for training, and V for validation. C means the Number of Classes, and SN means sensitivity.

#### 4.1.1. Limitations of the reviewed studies

1. Data Quality and Availability: There is a scarcity of high-quality, annotated datasets for both image Mezina and Burget (2024) and text Muaad et al. (2024) (e.g., X-rays, CT scans, covid-19 text) and audio data (cough recordings) Chen et al. (2024). Many existing datasets are either too small Choudhary et al. (2023), biased Manchanda et al. (2022)or lack the necessary diversity Issahaku et al. (2024). Additionally, datasets often have an imbalance between COVID-19 positive and negative samples, which can lead to biased models that perform well on the majority class but poorly on the minority class Galán-Cuenca et al. (2024).

2. Variability in Data: There is significant variability in how COVID-19 manifests in different patients, making it challenging to develop a one-size-fits-all model Rajaraman et al. (2020). Cough sounds can be heavily influenced by background noise, recording quality, and the presence of other respiratory conditions, making it difficult to isolate the features specific to COVID-19 Ulukaya et al. (2023) Campana et al. (2023).

3. Model Generalisation: Deep learning models are prone to overfitting, especially when trained on limited data. This results in models that perform well on training data but fail to generalise to new, unseen data Vaid et al. (2020) peng2020empirical. Furthermore, models trained on data from a specific population or geographical region may not generalise well to other populations due to differences in demographics, healthcare practices, and disease prevalence Afshar et al. (2021).

4. Interpretability and Explainability: Deep learning models, particularly those using convolutional neural networks (CNNs) Chen et al. (2024) for image analysis and recurrent neural networks (RNNs) Hassan et al. (2020), Liang et al. (2021) or transformers Ulukaya et al. (2023) Campana et al. (2023) Ardakani et al. (2020) for audio analysis, are often seen as black boxes. This lack of interpretability makes it difficult to understand the decision-making process and to gain clinical trust Singh and Yow (2021). Finding the features (for example, specific patterns in X-rays Bayram and Eleyan (2022) or specific sound characteristics in coughs) that are most indicative of COVID-19 can also be hard, which makes it harder to explain model predictions.

5. Regulatory and Ethical Issues: Collecting and using medical data, especially audio recordings of coughs, raises significant privacy concerns. Ensuring compliance with regulations like GDPR and HIPAA is essential Bartenschlager et al. (2024). There is also a risk of introducing bias into the models if the training data is not representative of the entire population, which could lead to disparities in diagnostic performance across different demographic groups.

6. Integration with Clinical Workflow: Many deep learning models are developed and tested in research settings and lack extensive clinical validation Wu et al. (2024). Integrating these models into real-world clinical workflows requires rigorous testing to ensure that they provide reliable and actionable results. Healthcare professionals may be hesitant to adopt AI-based tools without clear evidence of their benefits and without adequate training on how to use and interpret the results.

Addressing these limitations requires a multifaceted approach, including developing larger and more diverse datasets, improving model interpretability, conducting robust clinical validation, and carefully considering ethical and regulatory issues.

Table 6: Comparative result analysis.

| Ref | Data type | Method | Data Size | C | V | T:V(%) | Result | Findings with novel contributions and advancements | Research Gaps |
|---|---|---|---|---|---|---|---|---|---|
| Mezina and Burget (2024) | Image (X-ray) | Vision Transfer | 30,805 | 9 | CV | 80:20 | A: 81.9% | - Improved accuracy detecting nine long-COVID-19 conditions with a modified loss function, surpassing SOTA accuracy. | - No specific digital tool for post-COVID-19 pulmonary changes. |
| Chen et al. (2024) | Speech (Cough) | CNN | 1324 | 2 | CV | 80:20 | A: 72.7% | - Increased data volume through augmentation and segmentation.<br>- Improved accuracy and robustness through the ensemble method. | - Depending solely on cough analysis. |
| Muaad et al. (2024) | Text | Deep ensemble transfer learning | 39,000 | 2 | CV | 80:20 | A: 93% | - Enhanced accuracy through ensemble deep transfer learning.<br>- Improved data set balance through text samples sourced from Twitter. | - Potential bias in Twitter-sourced data.<br>- Reliance on Automatically Labeled Data. |
| Choudhary et al. (2023) | X-ray Image | ResNet | 2482 | 2 | CV | 80:20 | A: 95.47% | - Facilitates automatic detection of COVID-19.<br>- Used ResNet34 for COVID-19 detection. | - Conventional model trained with small data.<br>- Did not show a real-time image prediction. |
| Al-Waisy et al. (2023) | X-ray Image | ResNetV3 HRNet | 600 | 2 | CV | 80:20 | A: 99.9% | - Improves classification by handling relevant features.<br>- Efficient COVID-19 detection using a CNN-based classifier. | - Works with small data.<br>- Applicability and objectives are weak. |
| Islam et al. (2023a) | CT-ray Image | ResNetV3 | 13,800 | 2,3 | CV | 80:20 | A: 97% | - Grad-CAM enhances interpretability and accuracy by highlighting relevant image regions for COVID-19 detection.<br>- Offering insights without sacrificing performance. | - Time complexity needs to be reduced in real-time analysis. |
| Khan et al. (2023) | X-ray Image | Light CNN | 380 | 2 | 5K-CV | 80:20 | A: 98.8% | - Novel light CNN model with watershed region-growing segmentation for efficient chest X-ray feature extraction. | - Comparative Performance is low when working with multi-class data. |
| Ulukaya et al. (2023) | Cough | Dense and convolution | 1321 | 3 &2 | CV | 80:10:10 | A: 90.4% & 61.5% | - Can capture different and relevant neural features.<br>- Provides cloud-based application to analyse COVID-19. | - Its preprocessing is costly comparatively. |
| Chatterjee et al. (2023) | Cough | D-Cov19Net | 920 | 2 | CV | 80:20 | A: 97.2% | - Fast, robust, and accurate COVID-19 detection for early diagnosis and scalability. | - Convolution operation takes more time to extract features. |
| Campana et al. (2023) | Sound | VGGish, YAMNET, and L3-Net | 13,447 | 2 | CV | 80:20 | A: 81% | - An efficient way of detecting COVID-19 with phone audio input.<br>- It provides effectiveness and robustness in the extraction of audio features. | - Time complexity is higher comparatively. |
| Contreras Hernández et al. (2023) | Text | Transfer BERT | 959 | 2 | CV | 60:40 | A: 97% | - Efficient approach for short text classification.<br>- Able to classify multilingual Twitter text of COVID-19. | - Struggles with complex sentences and semantic nuances. |
| Ahmed et al. (2023) | Text | Lexicon-based deep learning | 764,398 | 2 | CV | 80:20 | A: 99% | - Customizable medical text recognition for COVID using deep learning and dictionaries. Easy and automated. | - Time complexity is higher than other methods |
| Paul et al. (2023) | Text | RNN MLP CNN | 20000 | 2 | CV | 80:20 | F1: 69% | - Efficient bilingual model to analyse COVID-19 text.<br>- Improves classification with independent features. | - Need to improve its method to handle context and sarcasm meaning of text |
| Qorib et al. (2023) | Text | BERT Lexicon | 10560 | 4 | CV | 80:20 | A: 96.71% | - Efficient text mining for Emotional Reactions to COVID-19 Vaccination Tweets, with quick computation. | - Need to improve to handle long sentences with deep contextual meaning. |
| Chakraborty et al. (2022) | X-ray | ResNet | 10,040 | 3 | CV | 80:20 | A: 96.43%; S: 93.68% | - Efficient COVID-19 detection with fast processing. ResNet trains deep neural networks effectively. | - Less adaptable.<br>- Lack of validation. |

table continues

continue table

| Ref. | Data type | Method | Data Size | C | V | T:V(%) | Result | Findings with novel contributions and advancements | Research Gaps |
|------|-----------|--------|-----------|---|---|--------|--------|---------------------------------------------------|---------------|
| Bhattacharyya et al. (2022) | X-ray Image | - VGG-19 - CGAN | 3750 | 3 | CV | 80:20 | A: 96.6% | - Automatic Identification of COVID-19 with CNN. - Provides efficient synthetic sample generation with GAN. | - Overfitting problems. - Works on small data. |
| Aggarwal et al. (2022) | X-ray Image | Transfer Model | 959 | 4 | CV | 80:20 | A: 97% | - Automatic COVID-19 detection using transfer learning. - Minimising labelled dataset needs, expediting the process. | - Small set images. - Performance is weak computationally. |
| Islam et al. (2022b) | X-ray Image | DenseNet | 231 | 3 | CV | 80:20 | A: 96.49% | - Identifies COVID-19 with CNN, maximising feature reuse and parameter efficiency for better performance. | - Overfitting in performance. - Works on small data. |
| Aly et al. (2022) | Speech | ML, DL | 1604 | 7 | CV | 80:20 | A: 96% | - Identify COVID-19 from speech records using machine learning. - Efficient COVID-19 detection with multi-domain data handling. | - Did not present a graphical result. - Less adaptable for big data. |
| Pahar et al. (2022) | Cough | ResNet | Various | 2 | CV | 80:20 | A: 0.98% (Cough), 0.94% (Breath), 0.92% (Speech) | - Identify COVID-19 from audio data and its bottleneck features. - Enhanced performance and generalization. | - Comparatively weak analysis presented. |
| Biswas and Dash (2022) | Text (Tweet) | CNN-LSTM | NA | 2 | CV | 80:20 | A: 75.72% | - COVID-19 post-feeling analysis. | - Analysis should be improved technically. |
| Malla and Alphonse (2022) | Text (Tweet) | BERT, ML, DL | 10,5700 | 2 | CV | 80:20 | A: 98.93% (DL); A: 98.88% (ML) | - Twitter post based COVID-19. - Post-feelings analysis. | - Try different models except transfer learning. |
| Jain et al. (2021) | X-ray Image | InceptionV3 XceptionNet ResNet | 6432 | 3 | CV | 80:20 | P: 99%; R: 92%; F1: 95% | - Identify COVID-19 using CNN. | - Overfitting. - Small data. |
| Sitaula and Hossain (2021) | X-ray Image | Attention VGG | 8571 | 3 | CV | 80:20 | A: 79.82%, 85.43%, 87.49% | - Handles independent features with an attention mechanism. - Using attention to the VGG model for COVID-19 classification. | - Weak in data augmentation. |
| Ozyurt et al. (2021) | CT Image | DNN | 19646 | 2 | CV | 80:20 | A: 95.84% | - Dynamic specimens with pyramid features enhance object detection accuracy and spatial localisation while maintaining computational efficiency and adaptability. | - Complex analysis. - Small data. |
| Zebin and Rezvy (2021) | X-ray Image | VGG16 ResNet50 Efficient Net | 822 | 3 | 5Fold | 80:20 | A: 90.0%; A: 94.3%; 96.8% | - Visualise infected lung area during classification of COVID-19. - Efficient way to handle important image features. | - Time and space complexity is higher. |
| Ahsan et al. (2021) | CT X-Ray | CNN | 5090 | 2 | KFold | 80:20 | A: 98.36% | - Identify COVID-19 with minimal complexity. - Efficiently handle the fusion of features. | - Imbalance data. - Extract irrelevant features sometimes. |
| Ismael and Şengür (2021) | X-ray Image | CNN SVM | 380 | 2 | CV | 80:20 | A: 94.7% | - Use SVM and CNN to quickly classify COVID-19. Deep features with SVM outperform local descriptors. | - Computationally Weak model. - Small data. |
| Zhang et al. (2021a) | X-ray Image | Attention GAN | 100 | 2 | CV | 80:20 | S: 69.85%; P: 94.6% | - Dense with GAN augmentation performs well in classification and segmentation, focussing attention on relevant regions for improved accuracy and robustness. | - Small data. |
| Wang et al. (2021) | CT Image | FGCN | 320 | 2 | CV | 80:20 | A: 97.71% | - Handles context-aware features and relations for better COVID-19 classification, with deeper adaptability. - Improved performance with spatial and structural fusion. | - Not enough adaptability. |
| Yu et al. (2021) | X-ray Image | ResGNet | 296 | 2 | CV | 80:20 | A: 96.62%; S: 95.91%; SN: 97.33% | - High performance in feature extraction by ResNet and GCN. - Handle spatial and structural information for better outcomes. | - Very small data. |

table continues

continue table

| Ref. | Data type | Method | Data Size | C | V | T:V(%) | Result | Findings with novel contributions and advancements | Research Gaps |
|---|---|---|---|---|---|---|---|---|---|
| Afshar et al. (2020) | X-ray Image | Capsule Network | 13800 | 3 | CV | 80:20 | A: 95.7%; P: 95.8%; SN: 90% | - Utilise capsule-RNN to find COVID-19 images.<br>- Ability to capture hierarchical spatial features. | - Not suitable for large data.<br>- Training time is higher. |
| Mobiny et al. (2020) | CT Image | Capsule Network | 746 | 2 | CV | 80:20 | A: 87.6%; P: 84.3%; S: 85.2%; F1: 87.1% | - Capsule-based net efficiently classifies COVID-19 CT scan images with adaptability. | - Very small data. |
| Islam et al. (2020a) | X-ray Image | CNN LSTM | 40000 | 3 | CV | 80:20 | A: 99.4%; S: 99.2%; F1: 98.9% | - Optimal CNN LSTM model for COVID-19 automatic detection.<br>- Properly capture independent features. | - Method should be improved for multi-class. |
| Lella and Pja (2022) | Sound | DCNN | 1539 | 5 | CV | 80:20 | A: 95.45%; F1: 96.96% | - Utilises multi-featured channels, enhancing COVID-19 diagnostic accuracy. | - Complex model.<br>- Augmentation should improve. |
| Sait et al. (2021) | Sound Image | Inception | 4558 | 2 | CV | 80:20 | A: 80%; A: 99.66% | - Combining breathing sounds and chest X-ray images enhances accuracy, reducing diagnostic errors and improving COVID-19 detection reliability. | - Higher time complexity.<br>- Small data. |
| Pinkas et al. (2020) | Sound | Attention GRU | 92 | 2 | CV | 80:20 | A: 87% | - Employs self-attention for better COVID-19 classification.<br>- Implements pre-training, bootstrapping, and regularisation to tackle overfitting. | - Leads miss information by correlation. |
| Pahar et al. (2021) | Sound | SVM LSTM ResNet | 1079 | 2 | CV | 80:20 | A: 98%; A: 94% | - Classifies COVID-19 patient cough sounds using a mobile app and ResNet-LSTM. Global smartphone recordings provide rapid, scalable early detection initiatives. | - Data should be balanced and extended. |
| Wang et al. (2020b) | Sound | Attention BiGRU | 4218 | 2 | CV | 80:20 | R: 83.7% | - Integrating visible light and infrared imagery to classify COVID-19. | pre-processing is hard. |
| Brown et al. (2020) | Sound | VGG | 142 | 2 | 10Fold | 80:20 | A: 80%; P: 72%; R: 69% | - Automatic detection of COVID-19 of asthma and cough-based patient. | - Weak feature selection and analysis. |
| Laguarta et al. (2020b) | Sound | ResNet | 5300 | 2 | CV | 80:20 | SN: 98.5%; S: 94% | - Efficiently classify cough utilizing biometric features and DL. | - High time complexity. |
| Dastider et al. (2021) | Sound | CNN-LSTM | 38 | 2 | CV | 80:20 | A: 67%; S: 97%; SN: 76.8%; F1: 66.6% | - High COVID-19 detection accuracy from lung sounds. Captures spatial and temporal patterns, aiding timely interventions. | - Performance is biased toward labelling and preprocessing. |
| Hassan et al. (2020) | Sound | LSTM | 240 | 2 | CV | 80:20 | A: 97%; A: 98.2%; A: 84.4% | - Classifies COVID-19 using LSTM with cough, breath and voice.<br>- Improves voice test accuracy with broader datasets. | - Small dataset and lacking labelling class. |
| Paka et al. (2021) | Text Tweet | Attention BiLSTM | 16000 | 2 | CV | 80:20 | A: 95.4%; P: 94.6%; R: 96.1%; F1: 95.3% | - Find fake Twitter news with Attention-based BiLSTM.<br>- Cross-SEAN improves performance using labelled and unlabelled data, boosting accuracy and robustness. | - Weak to handle retweet posts. |
| Nugraheni et al. (2020) | Text News | CNN | 16833 | 3 | CV | 80:20 | A: 94.99%; F1: 70.33% | - Classify aggravation of short post of COVID-19 with CNN.<br>- Faster in computation. | - Very simple model. |
| Liang et al. (2021) | Text | BiLSTM | 10700 | 2 | CV | 80:20 | A: 98.59% | - Detects fake news with BiLSTM. Analyses complex relationships in medical imaging data. | - Very conventional model. |
| Che et al. (2021) | Text Clinical | Attention GCN | 1000 | 5 | 10Fold | 80:20 | A: 95.40%; P: 85.10% | - Use GCN, knowledge, and attention to adaptable drug repositioning, enhancing performance. | - Less adaptable Data.<br>- Labelling controls performance. |
| Islam et al. (2021b) | Twitter | CNN-LSTM | 3668 | 4 | CV | 90:10 | P: 71.84%; R: 70.98%; F1: 71.40% | - A fine-grained analysis of missing information.<br>- Can handle the context of long text. | - Data annotation should be improved. |

table continues

continue table

| Ref. | Data type | Method | Data Size | C | V | T:V(%) | Result | Findings with novel contributions and advancements | Research Gaps |
|---|---|---|---|---|---|---|---|---|---|
| Parashar et al. (2020) | Text | CNN-LSTM | NA | NA | CV | 80:20 | A: 90.83% | - Analyzes and forecasts COVID-19 using a Hybrid model. Manage long-range and independent features effectively. | - Data information is not clear. |
| Zhang et al. (2020) | Text Social | Attention BiLSTM CRF | 25392 | 2 | CV | 80:20 | A: 85.49%; R: 83.14%; F1: 48% | - Classifies Chinese COVID-19 textual sentiments with Interactive Multitask deep Learning, efficiently handling semantics. | - Features should be improved. The method needs to be more adaptable. |
| Raamkumar et al. (2020) | Text Social | GRU | 16752 | 4 | CV | 80:20 | S: 96%; SN: 94.3%; P: 90.9% | - Classify public behavior for social distancing. | - Lack of adaptability of performance. |
| Saood and Hatem (2021) | X-Ray | U-Net SegNet | 100 | 2 | CV | 72:10 | A: 95% | - Nicely used the fusion of two hybrid models to analyze COVID-19.<br>- Properly handles global and local features. | - Comparatively higher complexity.<br>-Train on very tiny data. |
| Allioui et al. (2022) | X-Ray | Reinforcement learning | NA | 2 | CV | 80:20 | A: 97.12%; SN: 79.97%; SP: 99.48%; P: 85.21%; F1: 83.01% | - Good unsupervised approach for image-based detection of COVID-19.<br>- More adaptable. | - Model operation should be simplified. |
| Oda et al. (2022) | X-Ray | 2D-3D CNN | 1288 | 2 | CV | 80:20 | A: 83.3% | - An efficient method for COVID-19 detection using Deep CNN.<br>- Automatic and adaptable approach. | - Feature extraction process is slightly costly. |

## 4.2. Analytical result with ablation study

### 4.2.1. Experimental setup

The methodology proposed involved leveraging various pre-trained models. In our experimental framework, we refined these models to showcase their efficacy. The model was evaluated using binary cross-validation, dividing the dataset into 80% for training and 20% for validation. The model was trained and evaluated iteratively using binary cross-validation, and the findings were aggregated. The model development process entailed a maximum of 10 epochs and 256 mini-batches, utilising the Adam Optimiser with a learning rate set to 0.005 to minimise loss (L). To mitigate overfitting during training, regularisation of L2 with a coefficient of $1e^{-5}$ and dropout with a probability of 0.25 were applied. The adaptation of various pre-trained models for COVID-19 image, speech, and text analysis was progressively configured using three distinct datasets. The implementation was carried out using Python and Google's TensorFlow deep learning framework. Binary cross-validation was executed on a computer equipped with 5 CPUs (Intel(R) 3.60 GHz), 16 GB RAM, and running Windows 10. Overall model parameters tuning are presented in the Table 14 as Appendix of this paper.

## 4.3. Overall performance from the experimental analysis

To check and ensure the performance of recently developed models, we manually implemented eleven models and obtained individual precision and other performance in training. We have an analytical analysis of three different datasets (text, speech, and image) using the ten most advanced deep learning methods.

In the speech-based COVID-19 analysis, the MobileNet model also scored the highest accuracy of 93.73%, and the CNN-RNN model got the lowest accuracy of 90.55%. Covid Cough implemented data available at the link[1]. These data contain positive and negative COVID-19 samples. The total amount of sample data is 1257(166 Unique samples). Here, Table 7 shows the comparative accuracy performance in this dataset (speech), and Figure 10 presents the overall performance graph.

In text-based COVID-19 analysis, the BiGRU model got a top accuracy of 99.89, but the GAN model got 95.59%. Covid text implemented data available at the link[2]. These data contain positive 17250 and negative 15150 in 7600 tweets for the COVID-19 response.

Here, Table 7 shows the comparative accuracy performance on this data set (text), and Figure 11 presents the overall performance graph.

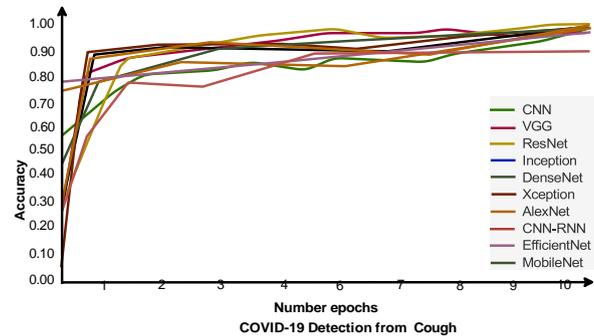

**Figure 10:** Performance of each model for cough-based COVID-19 recognition.

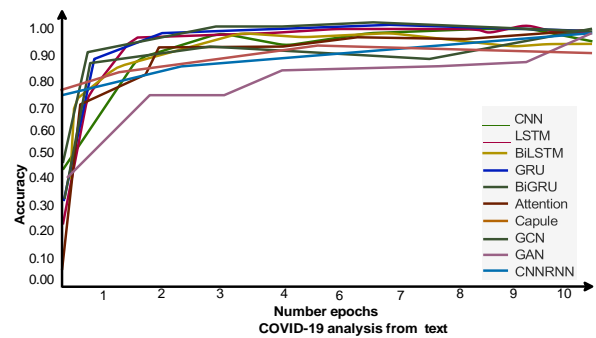

**Figure 11:** Performance of each model for Text-based COVID-19 analysis.

In implementing each model for image-based COVID-19 detection, we choose the X-ray image-based COVID-19 database verified by *Kaggle.com* of the author tawsifurrahman, this Covid image data available at the link[3]. We take 4000 X-ray images and 1000 images for four classes of image: normal, viral pneumonia, lung opacity, and COVID-19. Table 7 shows the training and validation accuracy of ten different deep learning models and loss during ten epochs. A graphical view of the performance (training and validation loss, accuracy) of eleven models is shown in Figure 12 for model CNN, VGG16, ResNet50, InceptionNetV3, NasNetMobile, DenseNet121, XceptionNet, AlexNet, CNN-RNN, EfficientNetB2 and MobileNetV2 both the train set and test set follow the same trend. Here, Table 7 and Figure 12 show the analysis of the critical results in COVID-19. MobileNetV2 is the best model with its highest training accuracy of 0.9797, and CNN based method performs as MobileNet with 0.9762 accuracy, and it gets the lowest training loss of 0.0367. In the validation performance, we see that VGG16 performs better with a validation accuracy of 94.20 and has the lowest validation loss of 0.09660. It is recommended that MobileNet performs as best for training and VGG16 performs as best for validation tasks for COVID-19 detection from X-ray images. In image-based COVID-19 detection, MobileNetV2 is the successful method with an accuracy of 97.97%, but ResNet obtained 77.31% accuracy.

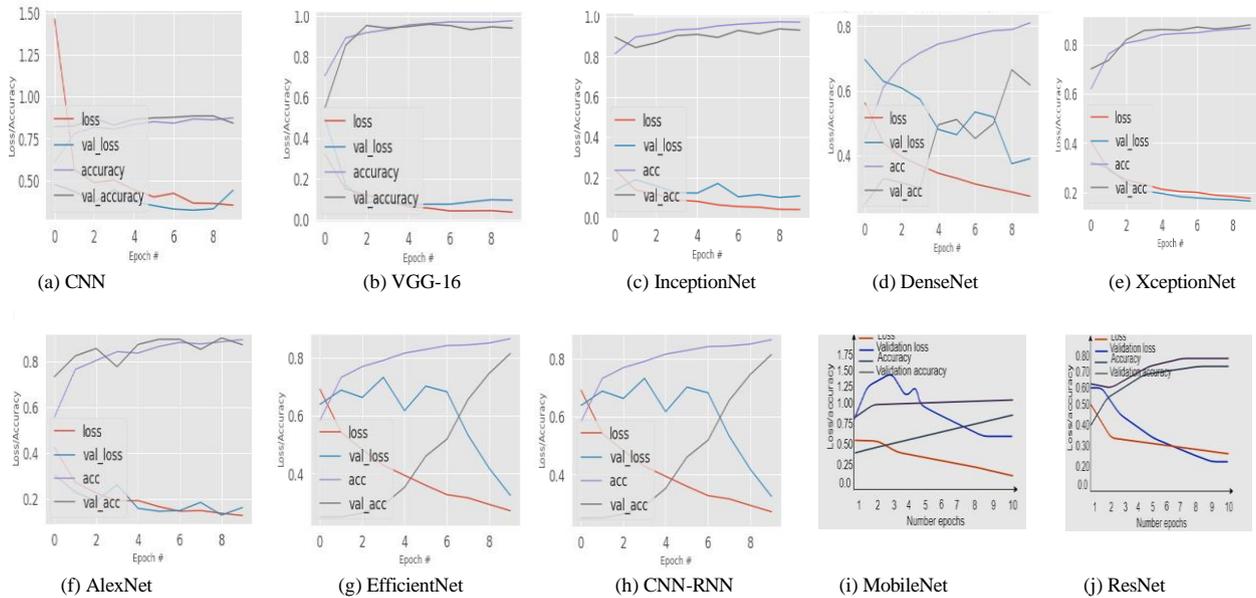

**Figure 12:** Performance of each model for Image-based COVID-19 recognition.

**Table 7**
Analytical analysis on three datasets(Train).

| COVID-19 Image analysis | | COVID-19 Speech analysis | | COVID-19 Text analysis | |
|---|---|---|---|---|---|
| Model | Accuracy | Model | Accuracy | Model | Accuracy |
| CNN | 97.01 | CNN | 92.96 | CNN | 89.50 |
| VGG-16 | 97.69 | VGG-16 | 93.44 | LSTM | 99.47 |
| ResNetV3 | 88.31 | ResNetV3 | 93.67 | BiLSTM | 99.63 |
| Inception | 97.06 | Inception | 92.95 | GRU | 99.85 |
| DenseNet | 81.01 | DenseNet | 92.84 | BiGRU | 99.89 |
| Xception | 86.62 | Xception | 93.22 | Attention | 98.34 |
| AlexNet | 89.66 | AlexNet | 93.68 | Capule | 98.77 |
| CNN-RNN | 86.62 | CNN-RNN | 90.55 | GCN | 96.67 |
| EfficentNet | 97.05 | EfficentNet | 92.53 | GAN | 95.59 |
| MobileNet | 97.97 | MobileNet | **93.73** | CNN-RNN | 96.84 |

## 4.4. Ablation study on COVID-19 test data (Image, cough and text)

Based on data provided in three different datasets (COVID-19 Image test data, COVID-19 Cough test data, and COVID-19 Text test data) and performance metrics of various models as shown in Table 8. In this table, A indicates accuracy, P indicates precision, F1 indicates F1 score, and S indicates support. Let us have a look at the analysed results:

1. **COVID-19 Image Test Data:** The models demonstrate strong performance on this test dataset(image), comprising 800 samples, achieving high accuracy scores ranging from 80.99% to 97.95%. In particular, the MobileNet model has the highest accuracy. In addition, its precision, recall, and F1 scores surpass those of other models, further underscoring its effectiveness. VGG-16 and Inception also achieve the highest accuracy scores, indicating their effectiveness in classifying COVID-19 images. DenseNet shows the lowest accuracy among the models, indicating potential limitations in its ability to extract features from COVID-19 image data.

2. **COVID-19 Cough Test Data:** The models show strong performance on this test dataset, comprising 166 samples, achieving high accuracy scores ranging from 90.52% to 97.94%. In particular, the MobileNet model has the highest accuracy. In addition, its precision, recall, and F1 scores surpass those of other models, further underscoring its effectiveness. Similarly, ResNet and AlexNet also show the highest accuracy scores, suggesting their effectiveness in analysing cough audio data associated with COVID-19. CNN-RN exhibits relatively lower accuracy compared to other models, indicating potential challenges in effectively capturing temporal dependencies in cough audio data.

**Table 8**
Analytical analysis on three datasets(Test).

| Model | COVID-19 Image test data | | | | | Model | COVID-19 Cough test data | | | | | Model | COVID-19 Text test data | | | | |
|---|---|---|---|---|---|---|---|---|---|---|---|---|---|---|---|---|---|
| | A | P | R | F1 | S | | A | P | R | F1 | S | | A | P | R | F1 | S |
| CNN | 96.98 | 96.97 | 96.94 | 96.97 | 800 | CNN | 92.92 | 92.91 | 92.92 | 92.92 | 33 | CNN | 89.49 | 89.45 | 89.46 | 89.43 | 8,000 |
| VGG-16 | 97.68 | 97.67 | 97.69 | 97.66 | 800 | VGG-16 | 93.39 | 93.37 | 93.36 | 93.38 | 33 | VGG-16 | 99.43 | 99.41 | 99.42 | 99.41 | 8,000 |
| ResNet V3 | 88.30 | 88.28 | 88.29 | 88.28 | 800 | ResNet V3 | 93.55 | 93.54 | 93.56 | 93.57 | 33 | BiLSTM | 99.59 | 99.54 | 99.55 | 99.53 | 8,000 |
| Inception | 97.04 | 97.03 | 97.02 | 97.02 | 800 | Inception | 92.78 | 92.77 | 92.78 | 92.6 | 33 | GRU | 99.84 | 99.81 | 99.82 | 99.80 | 8,000 |
| DenseNet | 80.99 | 80.98 | 80.97 | 80.99 | 800 | DenseNet | 92.81 | 92.80 | 92.81 | 92.79 | 33 | BiGRU | 99.87 | 99.85 | 99.86 | 99.82 | 8,000 |
| Xception | 86.55 | 86.54 | 86.55 | 86.54 | 800 | Xception | 93.17 | 93.14 | 93.15 | 93.13 | 33 | Attention | 98.31 | 98.27 | 98.28 | 98.27 | 8,000 |
| AlexNet | 89.65 | 89.64 | 89.65 | 89.63 | 800 | AlexNet | 93.65 | 93.62 | 93.63 | 93.61 | 33 | Capsule | 98.74 | 98.71 | 98.72 | 98.70 | 8,000 |
| CNN-RNN | 86.60 | 86.62 | 86.63 | 86.61 | 800 | CNN-RNN | 90.52 | 90.50 | 90.51 | 90.49 | 33 | GCN | 96.63 | 96.62 | 96.63 | 96.61 | 8,000 |
| EfficientNet | 97.02 | 97.01 | 97.02 | 97.01 | 800 | EfficientNet | 92.53 | 92.51 | 92.52 | 92.50 | 33 | GAN | 95.57 | 95.54 | 95.55 | 95.55 | 8,000 |
| MobileNet | 97.95 | 97.94 | 97.95 | 97.93 | 800 | MobileNet | 93.69 | 93.65 | 93.66 | 93.64 | 33 | CNN-RNN | 96.82 | 96.81 | 96.82 | 96.79 | 8,000 |

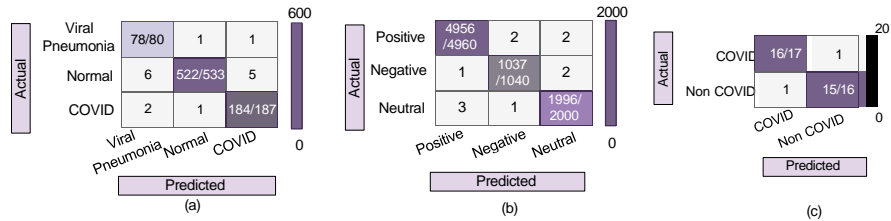

**Figure 13:** Confusion Matrix of best performed models utilising Image, Text and Cough test data.

3. **COVID-19 Text Test Data:** The performance on this dataset (with 8,000 samples) is also quite good, with accuracy scores ranging from 89.49% to 99.84%. GRU achieved the highest accuracy score, indicating their effectiveness in analysing textual data related to COVID-19. Similarly, LSTM, GRU, and BiLSTM models also obtained good accuracy. CNN and CNN-RNN show lower accuracy compared to other models, suggesting potential limitations in their ability to capture semantic features from textual data.

4. **Performance analysis with confusion matrix:** We have generated confusion matrices for the best performing model on the images, text, and cough-based data. In Fig13.a, the confusion matrix for the MobileNet model indicates that of the 187 samples in the COVID-19 class, 184 were correctly predicted using the test image data (800 samples), which also includes the classes Viral Pneumonia and Normal.
In Fig.13.b, the confusion matrix for the BiGRU model indicates that of the 4960 samples in the COVID-19 positive class, 4956 were correctly predicted using the test text data(8000 samples), which also includes the Negative and Neutral classes in the COVID-19 text.
In Fig.13.c, the confusion matrix for the MobileNet model indicates that of the 17 samples in the positive COVID-19 class, 16 were correctly predicted using the test cough data (33), which also includes the negative and neutral COVID-19 cough classes.

5. **Overall Observations:** Models such as MobileNet, Inception, BiGRU, and LSTM consistently perform well in different data sets, indicating their versatility in handling various types of COVID-19-related data (images, audio, and text). The choice of model architecture appears to play a significant role in the performance of different data modalities. For instance, recurrent neural networks (RNNs) and their variants demonstrate strong performance on text and audio data, while convolutional neural networks (CNNs) excel in image data classification tasks. The size of the data set varies significantly between different modalities, with the image data set having a sample size of 8,00, the text data set having the largest size (8,000 samples), and the cough dataset having the smallest size (33 samples). This difference in dataset size may impact the generalisation capability of models, with larger datasets generally leading to better performance.

## 4.5. Comparative result analysis of the experimental analysis with the state-of-the-art method

We summarised the comparative analysis of our methods with state-of-the-art techniques across different datasets:

For CT Image (Non and COVID-19) data analysis, Islam et al. (2023) achieved an accuracy of 97.00% using ResNetV3 on a dataset of 13,800 samples. Abdullah, Kedir, Takore et al. (2024) used DNN + SVM on a dataset of 19,412 samples, achieving an accuracy of 92.00%. Our MobileNet model achieved an accuracy of 97.97% on a dataset of 4,000 samples, surpassing previous methods. For X-ray (non-and COVID-19) data analysis, Islam et al. (2022b) reported an accuracy of 96.49% using DenseNet on a dataset of 231 samples.

For Cough (non- and COVID-19) data analysis: Ulukaya et al. (2023) achieved an accuracy of 90.40% with Dense and convolution methods on a dataset of 1,321 samples. Our MobileNet model attained an accuracy of 93.73% on a dataset of 166 samples, outperforming previous methods.

For COVID-19 Spread Text data analysis, Ahmed et al. (2023) used lexicon-based deep learning techniques, achieving an accuracy of 99.00% on a dataset of 764,398 samples. Our BiGRU model achieved a higher accuracy of 99.89% on a dataset of 40,000 samples, indicating superior performance. Fake News (COVID-19 Text): Akhter and Hossain (2024) reported an accuracy of 96.19% using CNN on a dataset of 3,120 samples.

**Table 9**
Comparative analysis with State of the art methods

| Author | Method | Dataset | Sample | Performance |
|---|---|---|---|---|
| Islam et al. (2023a) | ResNetV3 | CT Image (Non and COVID-19) | 13,800 | Accuracy 97.00% |
| Islam et al. (2022b) | DenseNet | X Ray (Non and COVID-19) | 231 | Accuracy 96.49% |
| Abdullah et al. (2024) | DNN+SVM | CT Image (Non and COVID-19) | 19,412 | Accuracy 92.00% |
| Ulukaya et al. (2023) | Dense and convolution | Cough (Non and COVID-19) | 1321 | Accuracy 90.40% |
| Campana et al. (2023) | VGGish, YAMNET, and L3-Net | Cough (Non and COVID-19) | 13,447 | Accuracy 81.00% |
| Kho and Shiong (2024) | ResNet | Cough (COVID-19) | 6,000 | Accuracy 92.10% |
| Contreras Hernández et al. (2023) | Mini VGGNet | Speech(Non and COVID-19) | 56 | Accuracy 92.00% |
| Pahar et al. (2022) | Transfer BERT | Text (Feedback on COVID-19) | 959 | Accuracy 97.00% |
| Ahmed et al. (2023) | Lexicon-based deep learning | Text (COVID-19 Spread text) | 764,398 | Accuracy 99.00% |
| Akhter and Hossain (2024) | CNN | Fake news (COVID-19 Text) | 3,120 | Accuracy 96.19% |
| **Our(Proposed)** | MobileNet | Image (X ray) | 4,000 | Accuracy 97.97% |
| **Our(Proposed)** | MobileNet | Speech(Cough) | 166 | Accuracy 93.73% |
| **Our(Proposed0** | BiGRU | Text (Comments) | 40,000 | Accuracy 99.89% |

**Table 10**
Feature list

| Feature list. | |
|---|---|
| F1: Attention | F2: Unsupervised |
| F3: Deep learning | F4: Hybrid |
| F5: NLP | F6: Capsule Network |
| F7: DNN | F8: CNN |
| F9: Speech | F10: Cross Validation |
| F11: MRI | F12: CT |
| F13: LSTM | F14: GRU |
| F15: RNN | F16: Machine learning |
| F17: GAN | F18: Embedding |
| F19: Neural Network | F20: GCN |

Our models consistently demonstrate competitive or superior performance across various datasets, showcasing their effectiveness in differentiating COVID-19 cases from non-COVID cases across multiple data modalities. Comparative result analysis of the experimental analysis with state of the art method are presented in the Table 9.

## 4.6. Quick summary of deep learning-based COVID-19 analysis techniques based on used features

A deep learning strategy for COVID-19 analysis uses a range of tools to forecast the outcome. In a deep learning model, different parameters are tuned, tools are embedded, pre-processing is done, and a supporting neural network with layers is used. In this part, we provide a quick summary Table 11 which highlights the facts about the technologies used in the deep learning-based COVID-19 analysis. Machine learning, as well as a supporting neural network model, have been used in hybrid or deep learning approaches to achieve effective outcomes. Many methods and characteristics are used in a deep learning-based COVID-19 analysis strategy. Several tools and features are often used in a deep learning model. We have put together a concise review of the most widely used deep learning-based COVID-19 analysis methods and characteristics. The twenty features listed in Table 10 were chosen based on the tools and techniques they use, as well as embedded and prepared data.

Here, Table 11 provides a concise summary of recent deep learning-based techniques for COVID-19 analysis. This table was created using our selection of twenty various features. We offer our most current work in this section, and we use the sign indicated to represent matched features across twenty features.

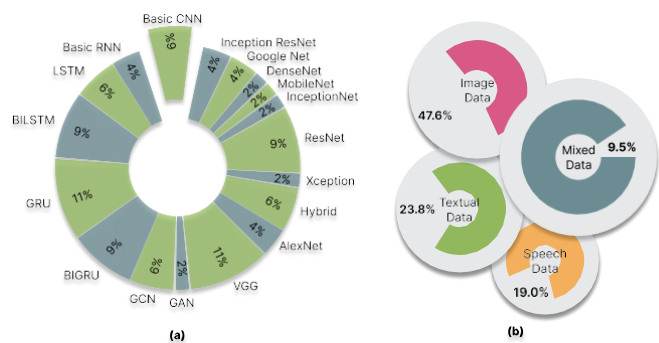

**Figure 14:** (a) Deep learning method amount for COVID-19 analysis, (b) Overview of the amount of data used COVID-19 analysis.

## 5. Discussion and Recommendation

### 5.1. Recommendation comparative analysis from literature

From the study of deep learning in this article, the deep learning algorithm has shown excellent results and important progress in the analysis of COVID-19 based on the image and speech of COVID-19. CNN and its hybrid method work better than the RNN-based method. However, in text-based COVID-19 analysis and prediction, RNN capsule-based methods with attention, capsules, and automatic encoder-decoders work better. Furthermore, in light of the COVID-19 pandemic, public health can be given priority with respect to personal privacy matters. To analyse COVID-19 with text, speech, or image data, the issues of privacy, human safety, and misuse of data should be highly considered for research transparency. In medical data analysis, government or political law, ethical concerns and privacy should be highly mentioned and considered. We contrasted the works mentioned on the basis of their transparency, implementation, and deep learning approaches to COVID-19 analysis. Figures 14a and 14b illustrate the number of different methods and data used in the COVID-19 analysis, respectively. In addition, we discuss possible directions for the study and problems related to the COVID-19 auxiliary data sets. The major obstacle to data-driven AI is the opacity of the data and the analysis methodology. The cough-based data set for COVID-19 diagnostics should be noise-free and small. Speech-based COVID-19 analysis requires more preprocessing time and noise-removing time. Most of the time, noisy data is the main obstacle to performance. Text-based data analysis of COVID-19 should be more realistic, large, and correctly labelled. Expanding CT scanning and X-ray data sets for greater precision of deep-learning technologies in image-based analysis.

**Table 11**
Overview of deep learning methods with selected features of tools used.

| Ref | F1 | F2 | F3 | F4 | F5 | F6 | F7 | F8 | F9 | F10 | F11 | F12 | F13 | F14 | F15 | F16 | F17 | F18 | F19 | F20 |
|---|---|---|---|---|---|---|---|---|---|---|---|---|---|---|---|---|---|---|---|---|
| Chakraborty et al. (2022) | | | √ | √ | | | | √ | | √ | | | | √ | | √ | | √ | √ | |
| Bhattacharyya et al. (2022) | | | √ | √ | | | | √ | | √ | | | | | | √ | | | √ | |
| Aggarwal et al. (2022) | | | | √ | | | | √ | | √ | | | | | | √ | | | √ | |
| Islam et al. (2022b) | | √ | √ | √ | | | | √ | | √ | | | | | | √ | | | √ | |
| Aly et al. (2022) | | | √ | √ | | | √ | √ | | | | | | | | √ | | √ | | |
| Pahar et al. (2022) | | | √ | √ | | | √ | √ | √ | | | | | | | √ | | | √ | |
| Biswas and Dash (2022) | √ | √ | √ | √ | | | | √ | | √ | | | √ | | | √ | | | √ | |
| Malla and Alphonse (2022) | | | √ | √ | | | | √ | | √ | | | | | | √ | | | √ | |
| Jain et al. (2021) | | | √ | √ | | | | √ | | √ | √ | | | √ | | √ | | √ | √ | |
| Sitaula and Hossain (2021) | √ | | √ | √ | | | | √ | | √ | | | | | | √ | | | √ | |
| Ozyurt et al. (2021) | | | √ | √ | | | √ | | | √ | | √ | | | | √ | | | √ | |
| Zebin and Rezvy (2021) | | | √ | √ | | | √ | √ | | √ | √ | | | | | √ | | | √ | |
| Ahsan et al. (2021) | | | | | | | | √ | | √ | √ | √ | | | | √ | | | √ | |
| Ismael and Şengür (2021) | | | √ | √ | | | | √ | | √ | √ | | | | | √ | | | √ | |
| Zhang et al. (2021a) | √ | | √ | √ | | | | | | √ | √ | | | | | √ | √ | | √ | |
| Wang et al. (2021) | | | √ | √ | | | | √ | | √ | √ | | √ | | | √ | | | √ | √ |
| Yu et al. (2021) | | | √ | √ | | | | √ | | √ | √ | | | √ | | √ | | √ | √ | |
| Afshar et al. (2020) | | | √ | √ | √ | √ | | √ | | √ | √ | | √ | | | √ | | | √ | |
| Prabha and Rathipriya (2020) | | | √ | √ | √ | √ | | √ | | √ | | √ | | | | √ | | | √ | |
| Islam et al. (2020a) | | | √ | √ | | | √ | √ | √ | √ | √ | | | | | √ | | | √ | |
| Parashar et al. (2020) | | | √ | √ | √ | | √ | √ | √ | √ | | | | | √ | √ | | | √ | |
| Zhang et al. (2020) | √ | | √ | √ | √ | | √ | √ | √ | √ | | | | | | √ | | | √ | |
| Raamkumar et al. (2020) | | | √ | √ | √ | | √ | √ | √ | √ | | | | | | √ | | | √ | |

Image-based data consists of more relevant information to detect COVID-19. However, almost all models are not applied enough to prove their real-world service, but they are already up to the mark in combating the SARS-CoV-2 outbreak. Our study has some limitations; a few specific aspects of the revised neural networks are not stated, especially for individual architectures, such as the number of layers, layer configuration, learning rate, number of periods, batch size, dropout layers, optimisers, and loss function. Although COVID-19 diagnosis is discussed from a computer-view point of view, this paper gives no qualitative diagnostic findings in CT or X-ray pictures, speech, and textual data. There are some disadvantages as well. To ensure a deep machine achieves the required output, deep learning techniques allow some labelled data to be trained. For COVID-19 research, we need big data to train the deep learning architecture to predict class labels correctly. Collecting and labelling a huge amount of data can be extremely complex and cumbersome. Deep learning works as a black box; if possible, adding knowledge to the model makes work unique hyperparameter tuning controls performance. Time and space complexity also greatly affect results, and deep learning needs high-performance-based hardware such as GPUs and wide RAM. From the review of different methods, we can recommend ResNet as the best model for picture data because it has 98% accuracy. We recommend the LSTM-based technique as the best-performing model for text-based COVID-19 analysis. The ResNet50-based model outperforms the other model in speech-based COVID-19 identification. In the case of our manual implementation of different deep learning methods, the recommendation is slightly different.

## 5.2. Experimental Recommendation

In our investigation, the LSTM-based technique achieved 98% accuracy in text-based COVID-19 analysis, whereas GRU achieved 96% accuracy. In speech-based COVID-19 recognition, the ResNet method works better with 98% accuracy, but LSTM is used more commonly. However, in text-based COVID-19 analysis, the BiGRU model got the highest accuracy of 99.89, but the GAN model got 95.59%. In image-based COVID-19 detection, MobileNetV2 is the successful method with an accuracy of 97.97%, but ResNet obtained 77.31% accuracy. In the speech-based COVID-19 analysis, the MobileNet model also scored the highest accuracy of 93.73%, and the CNN-RNN model got the lowest accuracy of 90.55%.

Most models have not had enough real-world testing, yet they can still counter the SARS-CoV-2 outbreak. In text-based COVID-19 analysis, the LSTM-based technique achieved 98 % accuracy, whereas GRU achieved 96% accuracy. However, while the ResNet method performs better with 98% in COVID-19 voice recognition, the LSTM method is more widely employed with its accuracy of 96%. Problem solving is a current task that potential researchers can do. Multimodal COVID-19 analysis is very rare Sait et al. (2021), Mahalle et al. (2020); this is one of the future works in COVID-19 research. From our analytical investigation of the most advanced ten deep learning model, it is recommended that MobileNet performs best for training and VGG16 performs best for validation tasks. Table 12 gives the advantages and limitations of different deep learning models for COVID-19 analysis of text, speech, and image data. Table 13 clearly shows the suggestion of various deep learning tasks based on the COVID-19 study.

## 5.3. Propose framework to develop a Cloud-based multimodal deep learning model

We have a potential plan to develop a Cloud-based multimodal deep learning model to analyse COVID-19 from speech, image, and text data. The future proposed framework is shown in Figure 15. In this model, we will input clinical COVID-19 data from different sources (public or private) and then preprocess it using cloud-based data pre-processing tools. After completing the preprocessing tasks successfully, a cloud-based deep learning model will extract features from the preprocessed data. The proposed framework will classify deep learning features with a machine learning algorithm. The combination of deep learning and machine learning classifier makes our model multimodal. Finally, the cloud-based result will be sent to the clinical scientist for delivery.

## 5.4. Standard Recommendation

To enhance its usefulness, the researcher could offer detailed, standard suggestions in the following areas:

1. **Standardised procedures to collect, curate and pre-process high-quality data sets for COVID-19 diagnosis.** Start by sourcing data from reliable public health databases and hospitals, ensuring ethical approvals and patient consent for COVID-19 diagnosis Elfer et al. (2024). Use de-identification algorithms to anonymize data and verify accuracy across sources. Engage medical experts for labeling, create an annotation manual, review regularly, and apply imputation, normalization, and statistical methods. Maintain

**Table 12**

Advantages and limitations of different deep learning models for COVID-19 analysis from text, speech, and image data.

| Model | Advantages | Disadvantages |
|---|---|---|
| CNN | - Automatically handles important features Hemdan et al. (2020), Ardakani et al. (2020).<br>- Little dependence on pre-processing Xu et al. (2020).<br>- Higher accuracy Ardakani et al. (2020).<br>- Fewer parameters Islam and Matin (2020), Chakraborty et al. (2022). | - Can not handle rotation of object Islam et al. (2020b).<br>- Slower than another model Islam and Matin (2020).<br>- Need proper initialisation Yu et al. (2021), Bhattacharyya et al. (2022). |
| RNN | - Can handle sequence data Jain et al. (2021).<br>- Can handle long-ranged features Hassan et al. (2020), Liang et al. (2021). | - Vanishing of the gradient Imran et al. (2020).<br>- Unable to handle long sequence with ReLU Raamkumar et al. (2020), Hassan et al. (2020). |
| LSTM | - LSTM has a memory facility.<br>- Handles features properly.<br>- Context handling properly.<br>- Dropout is hard to implement Hassan et al. (2020). | - Requires more memory.<br>- Overfitting problem. |
| GRU | - Faster than LSTM.<br>- Less time and space complexity.<br>- Good to handle long sequence Wang et al. (2020b). | - Slow convergence rate Wang et al. (2020b).<br>- Low learning efficiency.<br>- Takes a long training time Jiang et al. (2020).<br>- Underfitting problem Wang et al. (2020b). |
| CNN-RNN | - Works better to handle proper features.<br>- Works better for speech analysis Dastider et al. (2021), Ardakani et al. (2020). | Low performance in handling small objects Islam et al. (2020b), Ardakani et al. (2020). |
| Attention | - Easily handle target-oriented features Zhang et al. (2020).<br>- Deals with context and semantic Paka et al. (2021). | - Use more weight parameters Paka et al. (2021).<br>- Increased time complexity Pinkas et al. (2020). |
| Capsule | - Handles features from different angles.<br>- Suitable for detecting overlapping objects Malla and Alphonse (2021). | - Slow in computation Afshar et al. (2020).<br>- Higher complexity than CNN Afshar et al. (2020). |
| GAN | - Faster than CNN.<br>- Realistic in operation.<br>- No need for more preprocessing Fan et al. (2020), Elzeki et al. (2021), Goel et al. (2021). | - Training time is longer.<br>- Cannot handle complex input Quilodrán-Casas et al. (2022). |
| GCN | - Can handle useful features Liang et al. (2021).<br>- utilised graph advantage Yu et al. (2021), Kapoor et al. (2020). | - Operation time is longer Wang et al. (2021), Liang et al. (2021). |
| Hybrid | - Improve performance Shah et al. (2021).<br>- Deals with proper features Pustokhin et al. (2020), Biswas and Dash (2022). | - Operational complexity Saood and Hatem (2021).<br>- Structural complexity Mobiny et al. (2020). |

**Table 13**

Recommendation of tasks in deep learning based COVID-19 analysis.

| Issue | Observation | Recommendation |
|---|---|---|
| Data collection | Data from public or self-generation. | - Validated & augmented is highly needed. |
| Run System | CPU for normal operation. | - Cloud-based GPU is considered as best. |
| Data pre-processing | Prepressing helps to deal with irrelevant features for better analysis. | - Denoising and segmentation for image and speech-based analysis Pinkas et al. (2020), Ardakani et al. (2020).<br>- Irrelevant word removal is highly suggested. |
| Feature extraction | Good feature extraction from data gives better analysis. | - Encoder or transfer learning for image Ardakani et al. (2020).<br>- MFCC Brown et al. (2020), Khriji et al. (2021) for speech and BoW for text.<br>- Feature extraction is highly suggested. |
| Deep learning model | Various methods are used to analyse COVID-19 such as CNN, RNN, Attention, and Caps Net. | - RNN with Attention or Capsule is highly suggested Mobiny et al. (2020), Ardakani et al. (2020), Pustokhin et al. (2020). |
| Optimisation | Optimisation reduces the overall loss and increases accuracy. | - Adam Optimizer is highly recommended. |
| Evaluation metrics | The evaluation metric gives a performance score. | - Accuracy is popularly used |

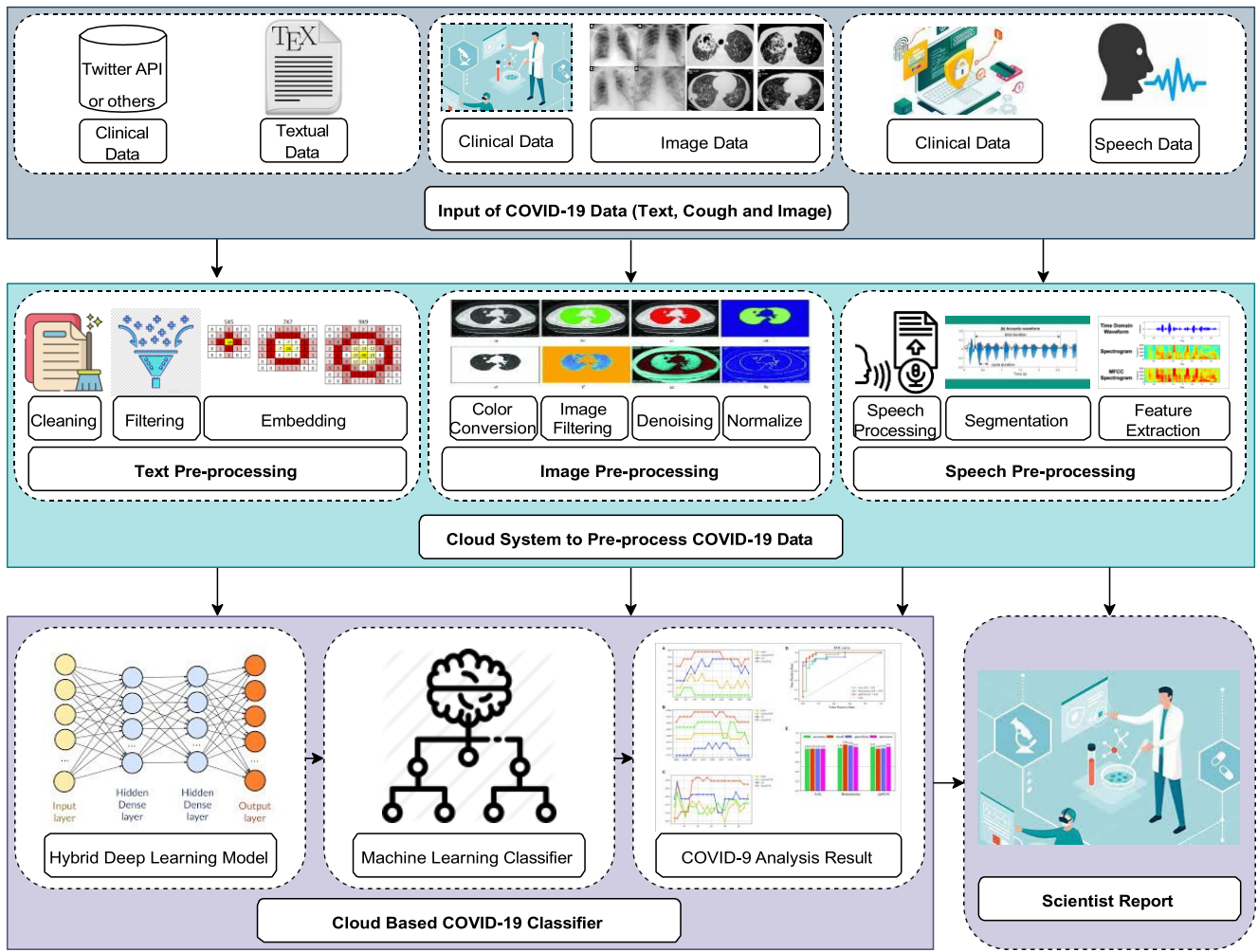

**Figure 15:** Cloud-based multi-modal deep learning Model to analyse multitype COVID-19 data.

records and seek community feedback. protocol Kumar et al. (2024).

(a) Development of Data Collection Protocols: Establish standardized guidelines for the collection of medical data, ensuring uniformity in COVID-19 data acquisition across different healthcare institutions Ealy et al. (2020). This would include criteria for patient inclusion/exclusion, data labelling, and anonymization processes to ensure privacy and ethical considerations Dou et al. (2021).

(b) Data Curation and Preprocessing Pipelines: Create robust frameworks for curating and preprocessing raw medical data Subrahmanian et al. (2023). This includes techniques for dealing with noise, missing values, and balancing datasets Akter et al. (2022) (e.g., for imbalanced classes like COVID-19 positive vs. negative). Implement methods such as augmentation or synthetic data generation to expand small datasets and maintain consistency in preprocessing steps.

(c) Collaboration with Multinational Institutions: Encourage collaborations across countries and organizations to gather diverse, multi modal datasets, enabling broader model generalization Rahman et al. (2021). Initiatives such as open-source repositories can facilitate data sharing and increase the dataset quality for COVID-19 and other related diseases Daramola et al. (2021).

2. **Integration of Latest Deep Learning Models:** To enhance diagnostic models with the latest deep learning, use state-of-the-art algorithms like transformers and graph neural networks. Employ transfer learning to fine-tune pre-trained models on COVID-19 datasets, and apply ensemble methods for improved robustness Ulukaya et al. (2023) Campana et al. (2023). Incorporate attention mechanisms for better interpretability and regularly update models with ongoing research.

(a) Adoption of Cutting-Edge Architectures Moreira et al. (2023): Incorporate the latest advancements in deep learning, such as transformers Mondal et al. (2021), attention-based models Zhou et al. (2021), and self-supervised learning techniques Li et al. (2020), into the existing COVID-19 diagnostic frameworks. These models can improve accuracy and interpretability Shome et al. (2021) in image-based or text-based diagnosis.

(b) Transfer Learning and Pretrained Models: Encourage the use of transfer learning with pretrained models, fine-tuned to COVID-19 diagnosis El Gannour et al. (2021). This would allow researchers to leverage large, publicly available medical datasets (e.g., chest X-rays, CT scans, cough and text) while focusing on COVID-19 specifics during training.

(c) Model Explainability and Interpretability: Prioritize the integration of explainable AI (XAI) techniques in deep learning models to ensure that healthcare practitioners can interpret model predictions effectively Ong et al. (2021). This could foster trust and reliability when deploying models in clinical environments Rodríguez et al. (2021).

3. **Combining Existing Models with Practical Applications:** To integrate models into healthcare settings and enhance diagnostic accuracy, establish strong collaboration between AI researchers and healthcare professionals for seamless implementation Sadeghi et al. (2024). Develop user-friendly interfaces and workflows that fit clinical practices, ensuring ease of use for medical staff Gheisari et al. (2024). Pilot the models in controlled environments to validate their performance and gather feedback for refinement Agarwal et al. (2024). Ensure robust data security and privacy measures to comply with regulations and build trust among stakeholders. Additionally, provide comprehensive training and support to healthcare providers to maximize the benefits of the integrated models Louati et al. (2024).

(a) Integration into Healthcare Workflows: Develop frameworks for embedding diagnostic models into hospital information systems and clinical workflows, especially in response to COVID-19 and similar pandemics Maghded et al. (2020). This may include real-time diagnostic assistance tools that integrate with electronic health records (EHRs) to provide clinicians with AI-generated insights while reviewing patient data, helping to track and manage infectious disease cases Bartlett et al. (2023).

(b) Cross-Modality and Hybrid Systems: Promote the use of multi-modal models Bian et al. (2022) that combine image-based, text-based, and sensor-based data, enhancing the diagnostic accuracy and robustness of AI systems, especially in COVID-19-related diagnostics. These models could integrate with wearable devices and real-time monitoring systems, aiding continuous health assessments for both hospital-bound patients and home-based care during pandemics Bayoudh (2023).

(c) Scalability and Deployment: Focus on creating scalable, cloud-based diagnostic solutions that can be easily deployed in various healthcare settings, particularly in low-resource environments Singh et al. (2022). This would help address healthcare disparities by enabling access to high-quality diagnostics globally, including for COVID-19 and future infectious disease outbreaks Hossain et al. (2020).

4. **Future Development Strategies:**
For future research in advancing diagnostics, explore integrating AI with wearable health monitoring devices to enable continuous health tracking and early disease detection Asif et al. (2024). Investigate federated learning techniques to train models using decentralized healthcare data while ensuring patient privacy. Examine the synergy between AI and genomics for personalized diagnostics and treatment based on genetic profiles. Utilize multimodal data fusion to combine imaging, clinical, and omics data for comprehensive insights Issahaku et al. (2024). Additionally, explore integrating AI with emerging technologies like quantum computing to address complex computational challenges in diagnostics Alejandro Lopez et al. (2024).

(a) Incorporation of Federated Learning: Leverage federated learning to allow AI models to learn from distributed COVID-19 datasets without compromising patient privacy Naz et al. (2022). This approach can help build more generalized models by aggregating knowledge from various hospitals without sharing sensitive COVID-19 data Zhang et al. (2021b).

(b) Application of Reinforcement Learning for COVID-19 Diagnosis: Explore the use of reinforcement learning (RL) Kumar et al. (2021) to optimize decision-making in diagnostics and treatment planning. RL could be used to adapt AI systems dynamically based on patient responses, leading to more personalized healthcare solutions Chen et al. (2022).

(c) Emerging Technologies to analyse COVID-19: Invest in the development and integration of emerging technologies such as quantum computing Sengupta and Srivastava (2021), which could significantly accelerate the training and deployment of deep learning models, particularly for large-scale diagnostic tasks Umer et al. (2022).

5. **Future Application Strategies in another field:** For future research in COVID-19 detection and other medical diagnoses, integrate advanced deep learning architectures like graph neural networks and reinforcement learning for better accuracy Alafif et al. (2021). Explore multi-modal data fusion combining clinical, imaging, and genetic data to enhance diagnostic capabilities Issahaku et al. (2024). Utilize transfer learning to adapt pre-trained models to new conditions, reducing the need for large labeled datasets. Consider AI integration with federated learning and edge computing for improved scalability and privacy Florescu et al. (2022). Promote interdisciplinary collaborations to accelerate innovation in diagnostics Bashir et al. (2024).

   (a) Recent advancements in Vision-Language Models (VLMs):

      Recent advancements in Vision-Language Models (VLMs) Van et al. (2024) offer significant potential for COVID-19 diagnosis by integrating imaging data (e.g., chest X-rays, CT scans) with clinical text (e.g., patient records, symptoms). VLMs can analyze visual and textual information simultaneously, providing more accurate and context-aware diagnostic insights.

      For example, in COVID-19 diagnosis, VLMs can combine the visual features of lung scans with clinical descriptions of symptoms and medical history to suggest a more accurate diagnosis or prognosis. This multimodal approach enhances diagnostic precision by factoring in both the visual anomalies typical of COVID-19 and patient-specific data such as pre-existing conditions. Moreover, VLMs can streamline the generation of comprehensive diagnostic reports by summarizing key findings from both image analysis and patient data Liu et al. (2021).

      These models can also adapt to evolving variants and symptoms by learning from continuous updates in clinical guidelines and imaging patterns, making them highly valuable in real-time pandemic management. Their ability to leverage multimodal data sources makes VLMs a promising tool for improving the accuracy and efficiency of COVID-19 and medical diagnostics Van et al. (2024).

Additionally, some other potential applications in COVID-19 analysis with deep learning span across multiple domains. Cardiac health monitoring has become crucial, as studies reveal long-term cardiac pathology even in individuals with mild initial illness Puntmann et al. (2022). Additionally, assessing lifestyle factors can help predict risks for post-COVID-19 complications, hospitalization, and mortality Wang et al. (2024). Vaccine safety research, such as the cardiovascular effects of various doses Ip et al. (2024) and socio-genetic predictors of vaccination status Hartonen et al. (2023), offers valuable insights for public health strategies. AI-driven diagnostic tools, including multi-modal image-text embedding and meta-learning models, are advancing medical diagnosis, though addressing bias concerns remains important DeGrave et al. (2021). Understanding national identity's role in public health support can inform effective communication strategies during global pandemics.

To enhance prediction accuracy, consider leveraging signal features like entropy and fractal dimension. Given the inherent ambiguity and uncertainty in medical data, employing fuzzy entropy and fuzzy regression proves beneficial. Fine-tuning with the improvement of the deep learning model may lead to better outcomes in COVID-19 diagnosis Talukder et al. (2024) Veluchamy et al. (2024) Ismail et al. (2024).

These aspects are intrinsically connected through their shared objective of enhancing diagnostic capabilities. High-quality data collection processes underpin the effectiveness of deep learning models, as these models require well-curated data to achieve accurate results. Integrating the latest deep learning advancements with existing models enhances diagnostic accuracy and efficiency, translating theoretical progress into practical healthcare tools. Future development strategies build on this foundation, exploring new technologies and methodologies to further improve diagnostics. The potential extends beyond COVID-19, as these advancements can be adapted to diagnose a variety of diseases, broadening the impact of innovative diagnostic approaches across multiple healthcare applications.

## 5.5. Limitation of the study

There are certain limitations concerning the application, including the data scarcity and vulnerability of the proposed method, particularly in its application and automation within clinical practice. The availability and quality of data, especially in the context of emerging diseases such as COVID-19, present significant challenges in the comprehensive analysis of its pre and post effects. Biases, inconsistencies, and limited data quantities could compromise the reliability and applicability of the developed models. Furthermore, considerations about generalisability, robustness, and ethical implications concerning patient privacy and data security deserve attention. Our focus has been primarily on the COVID-19 domain; however, to improve adaptability and usability, exploration of various domains with more in-depth analysis is imperative. Although we have not yet achieved automatic application with the proposed method, efforts toward its development are underway.

## 6. Conclusion

This work demonstrates that deep learning methods outperform other approaches in detecting and analyzing COVID-19, particularly in image-based diagnostics using CT and X-ray data, which have proven more effective than speech and text analyses. However, challenges remain, such as potential mispredictions in complex images and noise in speech data. Text data, in particular, require precise labelling and extensive training to achieve high performance. In this study, we presented a system designed using pre-trained models of deep transfer learning and customized deep-learning frameworks for diagnosing and analyzing COVID-19. Our three-tier taxonomy explores the potential of deep learning algorithms applied to images, text, and speech. We critically analyzed various advanced deep learning methods, discussing their operational details, equations, findings, applications, and performance. Moreover, we described the datasets used, outlined the challenges encountered, and recommended the best-performing models, aiming to provide useful insights to researchers entering this field.

However, it is important to note that while deep learning provides valuable insights, it does not replace the expertise of medical professionals in clinical diagnosis. The main challenges in deep learning for COVID-19 diagnostic systems include the lack of gold standards, dataset balancing, and labelling. However, collaboration between deep learning experts and medical professionals is essential to improve early diagnosis and assess the severity of infections. Future research focuses on creating more realistic and noise-free cough datasets and ensuring that text data are sufficiently large and accurately labelled. We anticipate further developments in real-time data analysis, multimodal analysis, and predictive medical assistance. Our future work will involve a more analytical review of models across different data types, leading to clearer performance assessments and recommendations.

## Acknowledgment

The authors would like to express their sincere gratitude to the anonymous reviewers for their valuable comments and insightful suggestions, which have significantly contributed to improving the quality and clarity of this article. Their constructive feedback has been instrumental in shaping the final version of this work.

**Table 14**

Models parameters tuning

| Model parameter tuning for Image-based COVID-19 Analysis | | | | | | | | | | | |
|---|---|---|---|---|---|---|---|---|---|---|---|
| Model | Total para- meters | Total sample | Train size | Preprocessing | Epoch | Batch Size | Target class | Learning rate | Optimiser | Activation function | Pooling | Drop out |
| CNN | 7M | 4000 | 3200 | Normalisation and Filtering | 10 | 256 | 4 | $1e^-5$ | Adam | ReLu | max pol | 0.25 |
| VGG-16 | 138M | 4000 | 3200 | Normalisation and Filtering | 10 | 256 | 4 | $1e^-5$ | Adam | ReLu | max pol | 0.25 |
| ResNet V3 | 25.6M | 4000 | 3200 | Normalisation and Filtering | 10 | 256 | 4 | $1e^-5$ | Adam | ReLu | max pol | 0.25 |
| Inception | 23.8M | 4000 | 3200 | Normalisation and Filtering | 10 | 256 | 4 | $1e^-5$ | Adam | ReLu | max pol | 0.25 |
| DenseNet | 8.06M | 4000 | 3200 | Normalisation and Filtering | 10 | 256 | 4 | $1e^-5$ | Adam | ReLu | max pol | 0.25 |
| Xception | 8.8M | 4000 | 3200 | Normalisation and Filtering | 10 | 256 | 4 | $1e^-5$ | Adam | ReLu | max pol | 0.25 |
| AlexNet | 60M | 4000 | 3200 | Normalisation and Filtering | 10 | 256 | 4 | $1e^-5$ | Adam | ReLu | max pol | 0.25 |
| CNN-RNN | 9.5M | 4000 | 3200 | Normalisation and Filtering | 10 | 256 | 4 | $1e^-5$ | Adam | ReLu | max pol | 0.25 |
| EfficientNet | 5.3M | 4000 | 3200 | Normalisation and Filtering | 10 | 256 | 4 | $1e^-5$ | Adam | ReLu | max pol | 0.25 |
| MobileNet | 4.25M | 4000 | 3200 | Normalisation and Filtering | 10 | 256 | 4 | $1e^-5$ | Adam | ReLu | max pol | 0.25 |
| **Model parameter tuning for Speech-based COVID-19 Analysis** | | | | | | | | | | | | |
| Model | Total para- meters | Total sample | Train size | Preprocessing | Epoch | Batch Size | Target class | Learning rate | Optimiser | Activation function | Pooling | Drop out |
| CNN | 7M | 166 | 132 | melSpectrogram | 10 | 256 | 2 | $1e^-5$ | Adam | ReLu | max pol | 0.25 |
| VGG-16 | 138M | 166 | 132 | melSpectrogram | 10 | 256 | 2 | $1e^-5$ | Adam | ReLu | max pol | 0.25 |
| ResNet V3 | 25.6M | 166 | 132 | melSpectrogram | 10 | 256 | 2 | $1e^-5$ | Adam | ReLu | max pol | 0.25 |
| Inception | 23.8M | 166 | 132 | melSpectrogram | 10 | 256 | 2 | $1e^-5$ | Adam | ReLu | max pol | 0.25 |
| DenseNet | 8.06M | 166 | 132 | melSpectrogram | 10 | 256 | 2 | $1e^-5$ | Adam | ReLu | max pol | 0.25 |
| Xception | 8.8M | 166 | 132 | melSpectrogram | 10 | 256 | 2 | $1e^-5$ | Adam | ReLu | max pol | 0.25 |
| AlexNet | 60M | 166 | 132 | melSpectrogram | 10 | 256 | 2 | $1e^-5$ | Adam | ReLu | max pol | 0.25 |
| CNN-RNN | 9.5M | 166 | 132 | melSpectrogram | 10 | 256 | 2 | $1e^-5$ | Adam | ReLu | max pol | 0.25 |
| EfficientNet | 5.3M | 166 | 132 | melSpectrogram | 10 | 256 | 2 | $1e^-5$ | Adam | ReLu | max pol | 0.25 |
| MobileNet | 4.25M | 166 | 132 | melSpectrogram | 10 | 256 | 2 | $1e^-5$ | Adam | ReLu | max pol | 0.25 |
| **Model parameter tuning for Text-based COVID-19 Analysis** | | | | | | | | | | | | |
| Model | Total para- meters | Total sample | Train size | Preprocessing and embedding | Epoch | Batch Size | Target class | Learning rate | Optimiser | Activation function | Pooling | Drop out |
| CNN | 7M | 40,000 | 32,000 | Basic Preprocessing, GloveVector Embedding | 10 | 256 | 3 | $1e^-5$ | Adam | ReLu | max pol | 0.25 |
| LSTM | 5.3M | 40,000 | 32,000 | Basic Preprocessing, GloveVector Embedding | 10 | 256 | 3 | $1e^-5$ | Adam | ReLu | max pol | 0.25 |
| BiLSTM | 7.1M | 40,000 | 32,000 | Basic Preprocessing, GloveVector Embedding | 10 | 256 | 3 | $1e^-5$ | Adam | ReLu | max pol | 0.25 |
| GRU | 7M | 40,000 | 32,000 | Basic Preprocessing, GloveVector Embedding | 10 | 256 | 3 | $1e^-5$ | Adam | ReLu | max pol | 0.25 |
| BiGRU | 7.1M | 40,000 | 32,000 | Basic Preprocessing, GloveVector Embedding | 10 | 256 | 3 | $1e^-5$ | Adam | ReLu | max pol | 0.25 |
| Attention | 5.3M | 40,000 | 32,000 | Basic Preprocessing, GloveVector Embedding | 10 | 256 | 3 | $1e^-5$ | Adam | ReLu | max pol | 0.25 |
| Capsule | 10.19M | 40,000 | 32,000 | Basic Preprocessing, GloveVector Embedding | 10 | 256 | 3 | $1e^-5$ | Adam | ReLu | max pol | 0.25 |
| GCN | 18m | 40,000 | 32,000 | Basic Preprocessing, GloveVector Embedding | 10 | 256 | 3 | $1e^-5$ | Adam | ReLu | max pol | 0.25 |
| GAN | 26.2M | 40,000 | 32,000 | Basic Preprocessing, GloveVector Embedding | 10 | 256 | 3 | $1e^-5$ | Adam | ReLu | max pol | 0.25 |
| CNN-RNN | 7M | 40,000 | 32,000 | Basic Preprocessing, GloveVector Embedding | 10 | 256 | 3 | $1e^-5$ | Adam | ReLu | max pol | 0.25 |

# Appendix A